\documentclass{article} 
\usepackage{iclr2027_conference,times}

\usepackage{amsmath,amsfonts,bm}

\def\eqref#1{equation~\ref{#1}}

\def\1{\bm{1}}

\DeclareMathAlphabet{\mathsfit}{\encodingdefault}{\sfdefault}{m}{sl}
\SetMathAlphabet{\mathsfit}{bold}{\encodingdefault}{\sfdefault}{bx}{n}

\usepackage{amsmath}
\usepackage{amssymb}
\usepackage{mathtools}
\usepackage{xcolor}
\usepackage{hyperref}
\usepackage{url}
\usepackage{adjustbox}
\usepackage{booktabs}
\usepackage{multirow}
\usepackage{array}
\usepackage{threeparttable}
\usepackage{cleveref}
\usepackage{float}
\usepackage{bbm}
\usepackage{wrapfig}
\usepackage{amsthm}
\newtheorem{proposition}{Proposition}
\newtheorem{corollary}{Corollary}[proposition]

\usepackage{adjustbox}

\AddToHook{cmd/appendix/before}{%
    \crefalias{section}{appendix}%
}
\crefname{appendix}{Appendix}{Appendices}

\newcommand{\gita}{\textsc{GITA}}
\newcommand{\hiql}{\textsc{HIQL}}
\newcommand{\ota}{\textsc{OTA}}
\newcommand{\gcbc}{\textsc{GCBC}}
\newcommand{\gcivl}{\textsc{GCIVL}}
\newcommand{\gciql}{\textsc{GCIQL}}
\newcommand{\qrl}{\textsc{QRL}}
\newcommand{\crl}{\textsc{CRL}}

\newcommand{\resultcell}[2]{%
    $#1${\scriptsize $\pm #2$}%
}

\newcommand{\resultcellbold}[2]{%
    $\mathbf{#1}${\scriptsize $\mathbf{\pm #2}$}%
}

\newcommand{\bestgammacell}[3]{%
    $#1${\scriptsize $\pm #2$}\,{\scriptsize $(#3)$}%
}

\newcommand{\factor}{\ensuremath{k}}

\title{Learning Multiple Timescales for\\ Goal-Conditioned Reinforcement Learning}

\author{%
\begin{tabular}{@{}ll@{}}
\textbf{Pedro Robles Dutenhefner}\textsuperscript{1}
&
\hspace{1.5em}
\textbf{Dikshant Shehmar}\textsuperscript{2,3}
\\[0.35em]
\textbf{Wagner Meira Jr.}\textsuperscript{1}
&
\hspace{1.5em}
\textbf{Marlos C. Machado}\textsuperscript{2,3,4}
\end{tabular}
\\[0.65em]
\textsuperscript{1}Universidade Federal de Minas Gerais,
\textsuperscript{2}University of Alberta
\\
\textsuperscript{3}Alberta Machine Intelligence Institute (Amii),
\textsuperscript{4}Canada CIFAR AI Chair
}
\begin{document}

\maketitle

\begin{abstract}
Existing approaches to offline goal-conditioned reinforcement learning (GCRL) struggle with long-horizon tasks. Discounting shrinks value differences between distant states until they fall below the function approximation error, leaving the agent with no signal for ranking states. Temporal abstraction---treating \(\factor\) environment steps as a single transition---restores this signal at long range, but no single fixed \(\factor\) suits all state-goal distances: large \(\factor\) preserves value differences across long temporal distances while collapsing distinctions between nearby states, and small \(\factor\) does the reverse. We make this trade-off explicit and introduce Generalized Implicit Temporal Abstraction (GITA), which conditions a single value function on \(\factor\). GITA trains one policy by aggregating advantage-weighted supervision across multiple \(\factor\) values, so scales assigning larger positive advantages to a state-goal pair contribute more strongly to its update. GITA does not need to choose between local resolution and long-range signal; it retains both without committing to a single \(\factor\). On OGBench, GITA outperforms a broad range of offline GCRL baselines, raising average success rate across all tasks by 25 percentage points (73\% relative improvement) over HIQL. It also improves over the strongest fixed-\(\factor\) method, OTA, by 7 percentage points (14\% relative).\footnote{Code is available at \url{https://github.com/pedroroblesduten/GITA}.}
\looseness=-1
\end{abstract}

\section{Introduction}
Offline goal-conditioned reinforcement learning (GCRL) aims to learn policies that reach arbitrary goals using only previously collected data \citep{levine2020offline}. They typically do so by learning a goal-conditioned value function \citep{schaul2015universal}. In long-horizon tasks, discounting shrinks the value difference between neighboring states as the goal moves farther away, and once that difference falls below the function approximator error, the learned values can no longer rank actions correctly \citep{opryshko2025test}. 

Temporal abstraction~\citep{sutton1999between} mitigates this issue by treating multiple primitive steps as a single transition. Even the simplest instantiation, fixing a stride of \(\factor\) steps, shortens the value-propagation chain for a given state-goal distance by a factor of \(\factor\). However, a large \(\factor\) that restores signal for distant goals also collapses distinctions between nearby states, as nearby states whose remaining distances fall within the same abstract-horizon bin become indistinguishable under the abstract transition. The right \(\factor\), therefore, depends on goal distance, and no fixed choice serves every state-goal pair. \looseness=-1


\begin{figure}[t]
\centering
\includegraphics[width=1.0\linewidth]{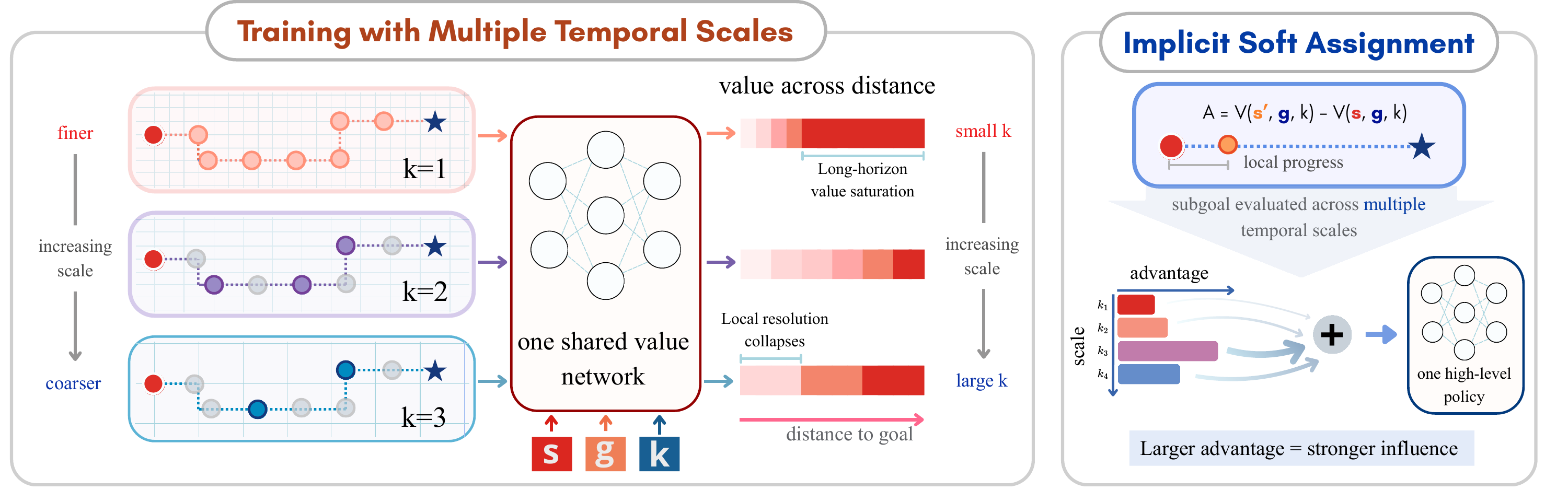}
\caption{\textbf{Overview of the proposed multi-scale temporal abstraction framework.} \textit{Left}: a shared value function conditioned on $k$ learns multiple abstraction scales. \textit{Right}: scale-specific advantages set the relative contribution of each $k$ value, inducing an implicit soft assignment when training the shared high-level policy. \looseness=-1}
\label{fig:diagram}
\end{figure}

In this paper, we introduce Generalized Implicit Temporal Abstraction (GITA), a GCRL method that learns a single \(\factor\)-conditioned value function spanning multiple temporal abstraction scales. From this value function, GITA derives an advantage at every scale and trains one policy on their aggregate via Advantage-Weighted Regression~\citep[AWR;][]{peng2019advantage}. Because AWR weights each sample by its exponentiated advantage, scales assigning larger positive advantages at a given state-goal pair contribute more strongly to the update. GITA therefore combines multiple effective horizons \emph{per state-goal pair} with no explicit selection mechanism---the sense in which the abstraction is implicit. It lifts the constraint that every fixed-scale method inherits by having nearby pairs resolved at small \(\factor\), distant ones at large \(\factor\), within one value function. \looseness=-1

Empirically, we characterize how \(\factor\) governs value discrimination and demonstrate GITA's effectiveness on OGBench~\citep{park2025ogbench}. In control environments where we can cleanly vary the distance between states, we show that distinguishing the values of two states at distance \(d\) requires \(\factor\) to grow with \(d\): below a threshold, the values of all sufficiently distant states collapse together. Conversely, a large enough \(\factor\) saturates the values of states closer than a threshold, collapsing those instead. No single \(\factor\) escapes both failures, which is precisely the trade-off GITA is designed to resolve. On OGBench, GITA outperforms existing hierarchical baselines, including HIQL~\citep{park2023hiql} and OTA~\citep{ahn2026option}, as well as a broad range of other GCRL methods across multiple morphologies, datasets, and environment types. In aggregate, GITA improves over HIQL by 25 percentage points (p.p.) of success rate (73\% relative) and over OTA by 7 p.p. (14\% relative). \looseness=-1

\section{Preliminaries}\label{sec:preliminaries}

We consider the offline GCRL setting~\citep{Kaelbling1993LearningTA, levine2020offline} defined over a Markov Decision Process (MDP), $\mathcal{M}=\langle\mathcal{S}, \mathcal{A}, P, \rho_0, r, \gamma, \mathcal{G}\rangle$. Here, $\mathcal{S}$ and $\mathcal{A}$ denote the state and action spaces respectively, $P(s' \mid s, a)$ is the transition dynamics, $\rho_0$ is the initial-state distribution, and $\gamma \in [0,1)$ is the discount factor. The goal space $\mathcal{G} \subseteq \mathcal{S}$ allows any valid state to serve as a goal~\citep{schaul2015universal, andrychowicz2017hindsight}, and $r(s,g)=-\mathbbm{1}[s\neq g]$ is a goal-conditioned reward function. We are given an offline dataset $\mathcal{D}$ of unlabeled trajectories $\tau=(s_0,a_0,s_1,\ldots,s_T)$ collected by an unknown behavior policy, and no further environment interaction is permitted. Writing $p^\pi(\tau\mid g)=\rho_0(s_0)\prod_{t}\pi(a_t\mid s_t,g)P(s_{t+1}\mid s_t,a_t)$ for the trajectory distribution induced by $\pi$, we want to learn a goal-conditioned policy that maximizes
$
    J(\pi)=\mathbb{E}_{g\sim p_{\mathcal{G}},\,\tau\sim p^\pi(\cdot\mid g)}\left[\sum_{t=0}^{H}\gamma^t r(s_t,g)\right],
$
where $p_{\mathcal{G}}$ and $H$ are the evaluation goal distribution and horizon. \looseness=-1

Hierarchical approaches, the main class of methods we study, facilitate long-horizon goal-conditioned control by introducing intermediate subgoals between the current state and the final goal. HIQL~\citep{park2023hiql} uses a high-level policy $\pi^h(s_{t+m}\mid s_t,g)$ to predict a subgoal $s_{t+m}$ at offset $m$, and a low-level policy $\pi^\ell(a_t\mid s_t,s_{t+m})$ to select primitive actions toward it. Both policies rely on a goal-conditioned value function trained with the expectile temporal-difference objective:
\begin{equation}
    \mathcal{L}_{V}
    =
    \mathbb{E}_{(s_t,s_{t+1})\sim\mathcal{D},\, g\sim p_{\mathcal{G}}}
    \left[
    L_2^\nu
    \left(
    r(s_t,g)
    +
    \gamma \bar{V}(s_{t+1},g)
    -
    V(s_t,g)
    \right)
    \right],
    \label{eq:hiql_value}
\end{equation}
where $s_{t+1}$ denotes the immediate successor state of $s_t$ in the dataset, $V$ is the goal-conditioned state-value function, $\bar{V}$ is a target network, and $L_2^\nu(u)=|\nu - \mathbbm{1}(u<0)|\,u^2$, with $\nu > 0.5$, denotes the expectile loss.

HIQL extracts both policies with AWR~\citep{peng2019advantage}. The high- and low-level advantages are
\begin{equation}
    \begin{gathered}
        A^h(s_t,s_{t+m},g)=V(s_{t+m},g)-V(s_t,g), \\
        A^\ell(s_t,s_{t+1},s_{t+m})=V(s_{t+1},s_{t+m})-V(s_t,s_{t+m}),
    \end{gathered}
\label{eq:hiql_advantages}
\end{equation}

giving the policy objectives
\begin{equation}
    \begin{gathered}
        J(\pi^h) = \mathbb{E}_{\mathcal{D},\, g\sim p_{\mathcal{G}}} \left[ \exp\!\left(\beta_h A^h\right) \log \pi^h(s_{t+m} \mid s_t,g) \right], \\
        J(\pi^\ell) = \mathbb{E}_{\mathcal{D}} \left[ \exp\!\left(\beta_\ell A^\ell\right) \log \pi^\ell(a_t \mid s_t,s_{t+m}) \right],
    \end{gathered}
\label{eq:hiql_policy_objectives}
\end{equation}
where $\beta_h,\beta_\ell > 0$ are inverse temperatures, and $A^\ell$ denotes the low-level advantage.

The high-level policy is trained from the advantage \(A^h = V(s_{t+m}, g) - V(s_t, g)\), which should be positive when the candidate subgoal \(s_{t+m}\) represents progress toward \(g\). Recovering this ordering requires the value function to resolve a difference between two states that are only \(m\) steps apart. As \(g\) moves farther away, one-step TD learning must propagate information over correspondingly longer horizons, and repeated discounting shrinks that difference until it is dominated by estimation error~\citep{ahn2026option}. The advantage then carries no reliable signal about which subgoals make progress, and high-level policy learning degrades.

Option-aware temporal abstraction~\citep[OTA;][]{ahn2026option} mitigates this issue by replacing the one-step value updates in Eq.~(\ref{eq:hiql_value}) with temporally-extended transitions. For a fixed abstraction factor $\factor$, OTA looks ahead $\factor$ primitive steps along the trajectory in $\mathcal{D}$ containing $s_t$ and uses $s_{t+\factor}$ as the successor state. Rather than forming a $\factor$-step TD target with accumulated rewards and discount $\gamma^\factor$, OTA treats this temporally-extended transition as a single update with TD error $\delta_\factor = r(s_{t+\factor},g)+\gamma \bar{V}^{h}_{\factor}(s_{t+\factor},g)-V^{h}_{\factor}(s_t,g)$, where $V^{h}_{\factor}$ denotes the temporally-extended value function. Each update therefore propagates value information across $\factor$ primitive steps while applying a single discount $\gamma$, reducing the number of backups needed to connect $s_t$ to $g$ from $d(s_t,g)$ to roughly $\lceil d(s_t,g)/\factor\rceil$, where $d(s_t, g)$ denotes the number of primitive steps from $s_t$ to $g$. Although a similar contraction can be induced by rescaling the discount as $\widetilde{\gamma}=\gamma^{1/\factor}$, \citet{ahn2026option} show that this alone does not reproduce the gains of temporally-extended value updates.

However, OTA fixes a single $\factor$ for all state-goal pairs while learning $V^h_\factor$. This makes $\factor$ a problem-dependent hyperparameter that must be tuned per environment, as \citet{ahn2026option} do. Moreover, since the number of backups separating $s_t$ from $g$ scales as $\lceil d(s_t,g)/\factor\rceil$, the $\factor$ that keeps this quantity small for distant goals is far larger than necessary for nearby ones, and states within $\factor$ steps become indistinguishable under $V^h_\factor$. 

We have focused here on the methods GITA builds on most directly. Appendix~\ref{sec:related_work} situates GITA within the broader offline GCRL and temporal abstraction literature.

\section{No Fixed Abstraction Suits All State-Goal Distances}
\label{sec:motivation}

We characterize how the temporal abstraction factor affects the high-level advantage used for subgoal selection. Let \(d=d^\star(s,g)\) denote the minimum number of primitive steps needed to go from a state \(s\) to a goal \(g\), and let \(s'\) be a candidate subgoal \(\Delta\) steps closer to the goal, such that \(d^\star(s',g)=d^\star(s,g)-\Delta\). For an abstraction factor \(\factor>0\), \(\factor\) primitive transitions are represented by one abstract transition, which incurs reward \(-1\) and a single discount \(\gamma\) under our analytical convention. Relaxing the abstract horizon \(\lceil d^\star(s,g)/\factor\rceil\) to \(d^\star(s,g)/\factor\) so that it varies smoothly with \(\factor\), as a continuous surrogate for the discrete-horizon problem, yields the temporally-abstract value function
\begin{equation}
\widetilde{V}_{\factor}(s,g)
=
-\frac{1-\gamma^{d^\star(s,g)/\factor}}{1-\gamma}.
\label{eq:temporally_abstract_value}
\end{equation}
Throughout, we assume \(d>\Delta>0\), \(\factor>0\), and \(0<\gamma<1\). Specifically, Proposition~\ref{proposition1} formalizes the relationship between the distance between different states and the difference between their induced value functions across different temporal scales \(k\). Proofs can be found in \cref{app:fixed_scale_tradeoff_derivation}. \looseness=-1

\begin{proposition}
Let \(s\) and \(s'\) denote two states such that \(s'\) is \(\Delta\) primitive steps closer to the goal than \(s\). When computed with a fixed stride of $k$ steps, the advantage of a subgoal \(s'\) in relation to \(s\) is \looseness=-1
\begin{equation}
\widetilde{A}_{\factor}(s,s',g)
=
\widetilde{V}_{\factor}(s',g)-\widetilde{V}_{\factor}(s,g)
=
\underbrace{\gamma^{(d-\Delta)/\factor}}_{\text{reachability}}
\;\cdot\;
\underbrace{\frac{1-\gamma^{\Delta/\factor}}{1-\gamma}}_{\text{resolution}}.
\label{eq:fixed_scale_advantage}
\end{equation}
\label{proposition1}
\end{proposition}

This proposition formalizes the intuition we have provided above. A coarser abstraction allows the advantage to better discriminate distant states when considering distant goals, because the \emph{reachability factor} increases toward $1$ as $\factor$ grows. However, the same coarser abstraction blurs the distinction between nearby states, since the \emph{resolution factor} decreases toward $0$ as $k$ grows. We formalize these results in the corollary below. \looseness=-1

\begin{corollary}
When computed with a fixed stride of $k$ steps, the advantage of a subgoal $s'$ in relation to $s$ vanishes either when the value of $k$ is too small or too large. Formally:
\begin{equation}
\lim_{\factor\to0^+}\widetilde{A}_{\factor}(s,s',g)=0,
\qquad
\lim_{\factor\to\infty}\widetilde{A}_{\factor}(s,s',g)=0.
\label{eq:fixed_scale_extreme_limits}
\end{equation}
\end{corollary}

Because \(d^\star(s,g)\), the minimum number of primitive steps needed to go from a state \(s\) to a
goal \(g\) varies depending on $s$, Proposition~\ref{proposition1} also allows us to conclude that no single $\factor$ can be equally effective in distinguishing between different pairs of states. We formalize this below.
\begin{corollary}
Let \(d = d^\star(s,g)\), and  \(\Delta = d^\star(s,g) - d^\star(s',g) \). For a state $s'$ where \(\Delta\ll d\), the optimal abstraction factor in the continuous relaxed model, $\factor_{\mathrm{opt}}^\star(d,\Delta)$, grows with the state-goal distance:
\begin{equation}
\factor_{\mathrm{opt}}^\star(d,\Delta)\approx-d\log\gamma.
\label{eq:fixed_scale_optimal_factor_local}
\end{equation}
This choice keeps the remaining state-goal distance approximately within one effective discount horizon:
\begin{equation}
\frac{d-\Delta}{\factor_{\mathrm{opt}}^\star}
\approx
-\frac{1}{\log\gamma}
\approx
\frac{1}{1-\gamma},
\qquad \gamma\to1.
\label{eq:fixed_scale_effective_horizon}
\end{equation}
\label{cor:optimal}
\end{corollary}

These predictions hold empirically. On a subset of OGBench environments, we train independent value functions \(V_{\factor}(s,g)\), one per \(\factor\), and evaluate them on the \((s,s',g)\) tuples used to train the high-level actor. Since subgoal selection depends only on the ordering induced by \(A_{\factor}(s,s',g)=V_{\factor}(s',g)-V_{\factor}(s,g)\), we measure the Spearman correlation between \(A_{\factor}\) and \(d(s,g)-d(s',g)\), which we call the Advantage-Progress Correlation (APC). We use geodesic shortest paths \citep{dijkstra2022note} as a proxy for \(d^\star\). See \Cref{app:fixed_scale_tradeoff} for more details. \looseness=-1 

\begin{figure}[t]
    \centering
    \includegraphics[width=1.0\linewidth]{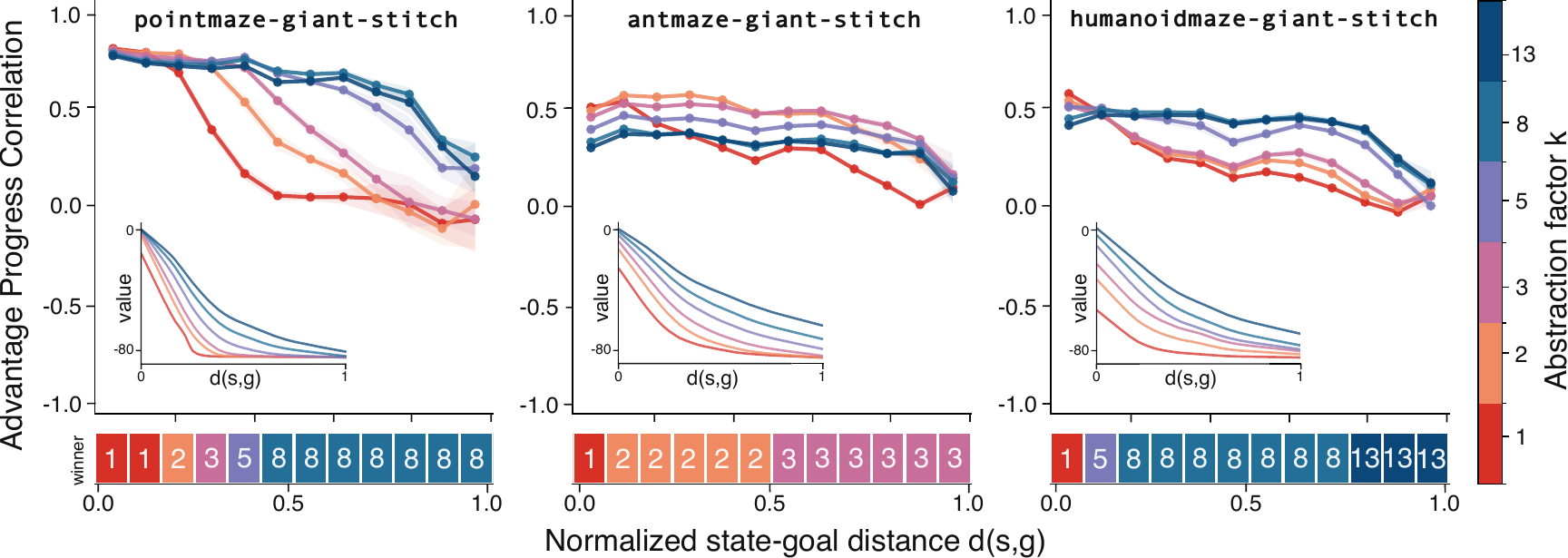}
    \vspace{-0.5cm}
    \caption{\textbf{No single abstraction factor is best at all distances.} APC for different abstraction factors \(\factor\) as the goal becomes more distant (mean \(\pm\) std over four seeds). The strip beneath each panel indicates the factor with the highest correlation in each distance regime; the best factor consistently grows with distance, but its range differs across environments. Inset shows the learned value \(V^h_{\factor}(s,g)\) as a function of geodesic distance. \looseness=-1}
    \label{fig:abstraction_factor_tradeoff_correlation}
\end{figure}

Figure~\ref{fig:abstraction_factor_tradeoff_correlation} shows a clear distance-dependent ordering across abstraction factors. Small factors achieve high APC at short state-goal distances but degrade rapidly as distance grows, consistent with the collapse of the reachability factor: in \texttt{pointmaze}, \(\factor=1\) falls to \(\text{APC}\approx0\) by \(d\approx0.4\), while \(\factor=8\) still retains \(\text{APC}\approx0.5\) at twice that distance. Larger factors bring distant pairs back into a value-sensitive range but attain lower APC at short distances, consistent with the loss of the resolution factor in Proposition~\ref{proposition1}. No factor dominates across the distance range; instead, the best-performing factor increases monotonically with state-goal distance in every environment, qualitatively consistent with the growth of \(\factor^\star_{\mathrm{opt}}\) with \(d\) predicted by Corollary~\ref{cor:optimal}. That range itself differs across environments---the best factor peaks at \(\factor=3\) in \texttt{antmaze} but \(\factor=13\) in \texttt{humanoidmaze}---so a single fixed choice cannot serve all environments. \looseness=-1

\section{Generalized Implicit Temporal Abstraction}
\label{sec:geo_ota}

Section~\ref{sec:motivation} established that, for locally separated states, the best abstraction factor grows with state-goal distance, so any fixed \(\factor\) is suboptimal for most pairs. \textbf{G}eneralized \textbf{I}mplicit \textbf{T}emporal \textbf{A}bstraction (GITA) removes the need to commit to any one \(\factor\). One value function conditioned on \(\factor\) represents all scales at once, and a single high-level policy is trained from the advantages this value function induces at every scale, weighted so that the scales best suited to each state-goal pair dominate its update. An overview of this mechanism is shown in Figure~\ref{fig:diagram}. The low-level policy is unchanged from HIQL, predicting primitive actions toward the subgoal the high-level policy proposes. \looseness=-1

\subsection{One Network, Multiple Abstraction Scales}
\label{sec:geo_ota_multiscale}

GITA introduces the abstraction factor \(\factor\) as an additional input to the high-level value network, learning a single parameter-shared \(V_\theta^h(s,g;\factor)\) where \(\factor\) is the number of primitive steps represented by each abstract transition. This makes the abstraction scale a controllable input rather than a fixed architectural choice: \(\factor\) is sampled during training, and \(V_\theta^h\) can be queried at any scale
afterwards. From here we write \(\factor\) as an explicit argument rather than a subscript to emphasize the conditioning. \looseness=-1

Given a trajectory \(\tau=(s_0,a_0,s_1,\ldots,s_T)\), let \(s_t\) denote the current state, \(g\sim p_{\mathcal{D}}\) a sampled goal relabeled either from a future state of \(\tau\) or from a random state in \(\mathcal{D}\), and \(s_{t+\factor}\) the state reached \(\factor\) primitive steps after \(s_t\) along \(\tau\), or the first state at which the goal is reached or the trajectory terminates, whichever comes first. For each training example, GITA independently samples an abstraction factor from a geometric distribution,
\begin{equation}
\factor \sim \operatorname{Geom}(1-\alpha), \qquad p(\factor)=(1-\alpha)\alpha^{\,\factor-1}, \quad \factor\geq 1,
\label{eq:geo_ota_geometric_sampling}
\end{equation}
where \(\alpha\in(0,1)\) sets the mean scale \(\mathbb{E}[\factor]=1/(1-\alpha)\). Geometric sampling favors shorter scales while retaining a long tail over larger ones, covering both fine and coarse temporal resolutions. \looseness=-1

GITA trains the abstraction-conditioned value function with the temporal-difference objective
\begin{equation}
\mathcal{L}_{V_\theta^h}
=
\mathbb{E}_{s_t\sim\mathcal{D},\,g\sim p_{\mathcal{D}},\,\factor\sim p(\factor)}
\left[
L_2^\nu
\left(
    r(s_{t+\factor},g)
    +
    \gamma \bar{V}^h_{\bar{\theta}}(s_{t+\factor},g;\factor)
    -
    V_\theta^h(s_t,g;\factor)
\right)
\right],
\label{eq:geo_ota_value}
\end{equation}
where \(\bar{V}^h_{\bar{\theta}}\) is a target network and the successor is evaluated at the same \(\factor\) that generated it, so each scale is trained against its own bootstrapped target. Since all scales are represented by a shared network, they are not learned independently, allowing information to generalize across scales. \looseness=-1

\begin{figure}[t]
    \centering
    \includegraphics[width=1.0\linewidth]{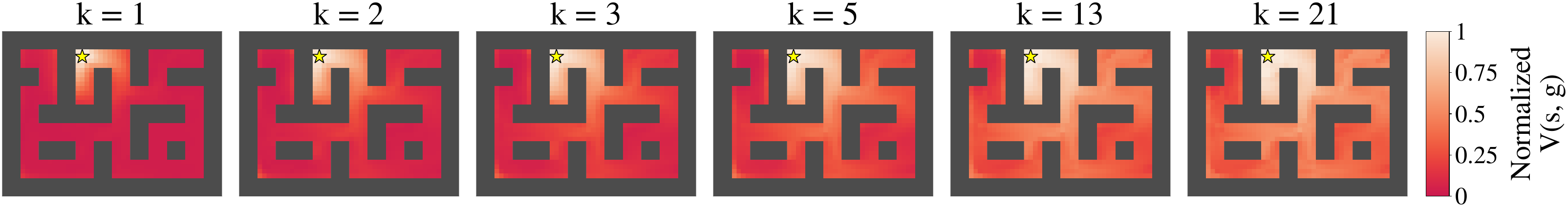}
    \vspace{-0.4cm}
    \caption{\textbf{Varying \(\factor\) directly controls the temporal abstraction scale.} Normalized value function \(V(s,g;\factor)\) on \texttt{pointmaze-large-navigate}, evaluated at every navigable cell for \(\factor \in \{1,2,3,5,13,21\}\). The star marks the fixed reference state. As \(\factor\) increases, the value-sensitive region expands over longer distances while local resolution progressively degrades.}
    \label{fig:value_function}
\end{figure}

\subsection{High-Level Policy Training with Multi-Scale Advantages}
\label{sec:geo_high_policy}

The high-level policy is trained by evaluating each training tuple under several temporal scales at once. We fix a finite set \(\mathcal{K}\) of abstraction factors as a hyperparameter and evaluate every tuple \((s_t,s_{t+m},g)\) under all \(\factor\in\mathcal{K}\), where \(m\) is the subgoal offset in primitive steps and is held fixed independently of \(\factor\). This is done so the candidate subgoal is the same, only the scale at which its advantage is measured varies. The policy itself is \emph{not} conditioned on the abstraction factor, giving a single \(\pi_\phi^h(s_{t+m}\mid s_t,g)\) shared across all scales.

For every tuple and every \(\factor\in\mathcal{K}\), we compute the abstraction-conditioned high-level advantage
\begin{equation}
A^h(s_t,s_{t+m},g;\factor)
=
V_\theta^h(s_{t+m},g;\factor)
-
V_\theta^h(s_t,g;\factor).
\label{eq:geo_high_advantage}
\end{equation}
Each abstraction scale therefore contributes a distinct evaluation of the same observed subgoal. We then aggregate the corresponding Advantage-Weighted Regression (AWR) weights across scales,
\begin{equation}
W(s_t,s_{t+m},g)
=
\frac{1}{|\mathcal{K}|}
\sum_{\factor\in\mathcal{K}}
\exp\!\left(
\beta_h A^h(s_t,s_{t+m},g;\factor)
\right),
\label{eq:geo_multiscale_weight}
\end{equation}
and train the high-level policy according to
\begin{equation}
\mathcal{L}_{\pi_\phi^h}
=
-
\mathbb{E}_{s_t,s_{t+m}\sim\mathcal{D},g\sim p_\mathcal{D}}
\left[
W(s_t,s_{t+m},g)
\log
\pi_\phi^h(s_{t+m}\mid s_t,g)
\right].
\label{eq:geo_high_actor}
\end{equation}

\begin{wrapfigure}{r}{0.45\textwidth}
    \centering
    \includegraphics[width=0.95\linewidth]{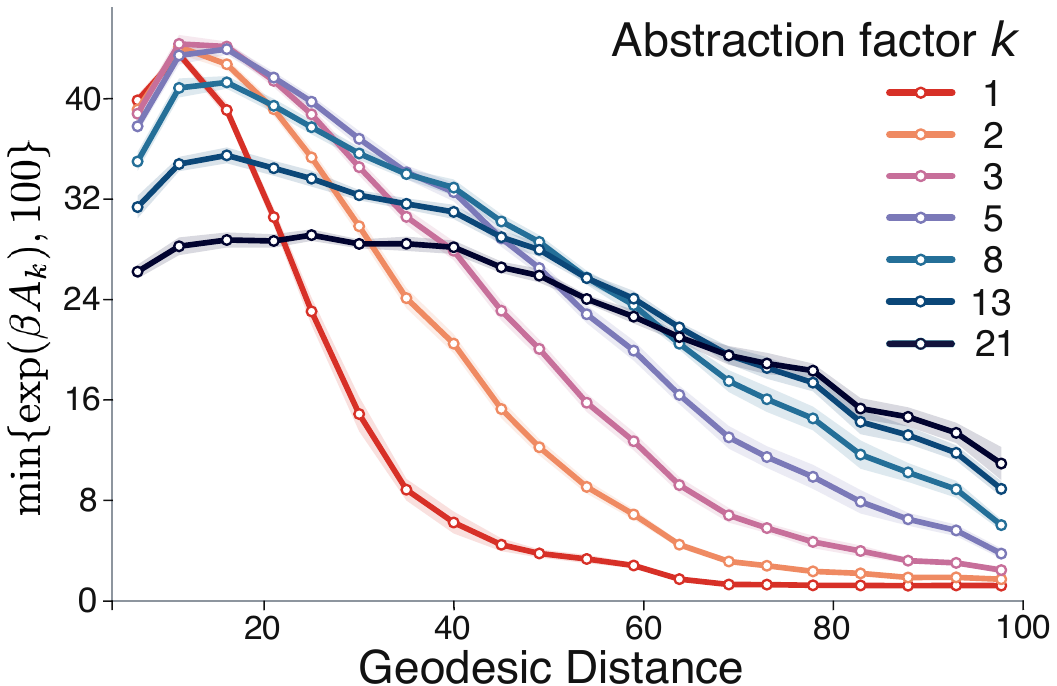}
    \caption{\textbf{The abstraction scale receiving the strongest policy weight changes with state-goal distance.} Clipped scale-specific AWR weight \(\min\{\exp(\beta_h A^h(s_t,s_{t+m},g;\factor)),100\}\) across geodesic state-goal distance bins in \texttt{antmaze-giant-navigate}, for each \(\factor\). \looseness=-1}
    \label{fig:adv_dist_giant_navigate}
    \vspace{-0.25cm}
\end{wrapfigure}

Applying the exponential before averaging makes \(W\) a soft aggregation of scale-specific AWR weights. Each term \(\exp(\beta_h A^h)\) is a weight in its own right, so scales distorted by saturation or excessive abstraction contribute weights near \(\exp(0)\), while the scale with the clearest positive advantage drives the update. By Proposition~\ref{proposition1}, that scale is the one nearest \(\factor_{\mathrm{opt}}^\star\), since factors below it are suppressed by the reachability factor and factors above it by the resolution factor. This is where the \emph{implicit} in GITA comes from: \Cref{sec:motivation} prescribes an abstraction factor that grows with \(d^\star(s,g)\), but \(d^\star\) is unavailable, and GITA never estimates it. Every scale in \(\mathcal{K}\) is placed on equal footing and the weighting resolves the competition on its own. \looseness=-1

Figure~\ref{fig:adv_dist_giant_navigate} shows the clipped scale-specific AWR weight against state-goal geodesic distance in \texttt{antmaze-giant-navigate}. At short distances, smaller abstraction factors receive the largest weights, but their advantage signal decays rapidly as the goal moves farther away, causing their weights to approach the neutral value \(\exp(0)=1\). Larger factors retain non-neutral weights over longer distances and become dominant at longer distances. This mirrors the distance-dependent ordering in Figure~\ref{fig:abstraction_factor_tradeoff_correlation}, which is what makes the implicit assignment work, as the scale that receives the strongest weight in a given distance regime is also the scale that ranks subgoals most reliably there. See \Cref{app:further_results} for additional environments.

At inference, the policy is executed directly as \(s_{t+m}\sim\pi_\phi^h(\cdot\mid s_t,g)\) and needs no abstraction factor as input. The low-level policy training objective is unchanged, with the policy predicting a primitive action conditioned on the current state and the generated subgoal, \(a_t\sim\pi_\psi^\ell(\cdot\mid s_t,s_{t+m})\).

\section{Experiments}

We evaluate GITA on the locomotion and manipulation tasks of the Offline Goal Conditioned RL Benchmark~\citep[OGBench;][]{park2025ogbench} suite, which includes long-horizon navigation tasks and datasets that require stitching together sub-trajectories to reach distant goals. We first describe our evaluation protocol, then compare GITA against representative offline GCRL approaches, including the best fixed-\(\factor\) alternative. \looseness=-1

\subsection{Evaluation on OGBench}

\paragraph{Experimental Setup.} We consider three locomotion tasks: PointMaze, AntMaze, and HumanoidMaze, controlling a 2-DoF ball, an 8-DoF ant, and a 21-DoF humanoid, respectively. Each comes with layouts of increasing geodesic distance---\texttt{medium}, \texttt{large}, and \texttt{giant}---and the \texttt{navigate} and \texttt{stitch} datasets; AntMaze additionally includes \texttt{explore}. We also consider two manipulation environments, Cube and Scene, where a robotic arm must reach a target configuration. The locomotion tasks probe long-horizon navigation, while the manipulation tasks probe compositional generalization. We use both the state- and pixel-based variants. An episode terminates when the agent comes within a task-specific proximity threshold of the goal state, which defines success, or when the environment's time limit is reached. Following the OGBench evaluation protocol, we report the average success rate across five goal-reaching tasks per environment, with $50$ evaluation episodes per task, averaged over $8$ random seeds for state-based tasks and $4$ for pixel-based tasks. \looseness=-1

We compare against the offline GCRL baselines provided by OGBench~\citep{park2025ogbench}: goal-conditioned behavior cloning~\citep[GCBC;][]{ghosh2019learning}, goal-conditioned implicit V- and Q-learning~\citep[GCIVL and GCIQL;][]{kostrikov2021offline,park2023hiql}, quasimetric RL~\citep[QRL;][]{wang2023optimal}, contrastive RL~\citep[CRL;][]{eysenbach2022contrastive}, hierarchical implicit Q-learning~\citep[HIQL;][]{park2023hiql}, and fixed-scale option-aware temporally abstracted value learning~\citep[OTA;][]{ahn2026option}. The comparison with OTA is the most informative, as it directly contrasts GITA's multi-scale formulation with fixed-scale temporal abstraction. HIQL abstracts only the policy, leaving its value function at \(\factor=1\), so the three together trace the progression from no value-level abstraction to fixed-scale to multi-scale. Note also that OTA tunes \(\factor\) per environment, whereas GITA uses a single candidate set \(\mathcal{K}\) throughout, avoiding per-environment tuning of the candidate set. See \cref{app:baselines} for further details about the baselines considered. \looseness=-1

\begin{table*}[t]
    \centering
    \begin{threeparttable}

        \caption{\textbf{GITA outperforms fixed-scale and non-abstracted baselines across OGBench tasks.}
        Success rate (\%), mean $\pm$ standard deviation over 8 seeds (4 for pixel-based tasks).
        The best mean, and any result within one standard deviation of it, are in \textbf{bold}.}
        \label{tab:all_baselines_ota_geo}

        \footnotesize
        \setlength{\tabcolsep}{2.0pt}
        \renewcommand{\arraystretch}{1.18}

        \begin{tabular}{
            @{}
            l
            l
            l
            c
            c
            c
            c
            c |
            c
            c
            c
            @{}
        }
            \toprule

            \multicolumn{3}{c}{}
            & \multicolumn{5}{c|}{\textbf{Non-Hierarchical}}
            & \multicolumn{3}{c}{\textbf{Hierarchical}} \\

            \cmidrule(lr){4-8}
            \cmidrule(lr){9-11}

            \textbf{Environment}
            & \textbf{Type}
            & \textbf{Size}
            & \textbf{\gcbc}
            & \textbf{\gcivl}
            & \textbf{\gciql}
            & \textbf{\qrl}
            & \textbf{\crl}
            & \textbf{\hiql}
            & \textbf{\ota}
            & \textbf{\gita} \\

            \midrule


            \multirow{4}{*}{PointMaze}
            & navigate
            & large
            & \resultcell{29}{6}
            & \resultcell{45}{5}
            & \resultcell{34}{3}
            & \resultcellbold{86}{9}
            & \resultcell{39}{7}
            & \resultcell{58}{5}
            & \resultcell{85}{5}
            & \resultcellbold{93}{2} \\

            & navigate
            & giant
            & \resultcell{1}{2}
            & \resultcell{0}{0}
            & \resultcell{0}{0}
            & \resultcell{68}{7}
            & \resultcell{27}{10}
            & \resultcell{46}{9}
            & \resultcell{72}{6}
            & \resultcellbold{80}{3} \\

            \cmidrule(lr){2-11}

            & stitch
            & large
            & \resultcell{7}{5}
            & \resultcell{12}{6}
            & \resultcell{31}{2}
            & \resultcellbold{84}{15}
            & \resultcell{0}{0}
            & \resultcell{13}{6}
            & \resultcell{46}{7}
            & \resultcellbold{72}{13} \\

            & stitch
            & giant
            & \resultcell{0}{0}
            & \resultcell{0}{0}
            & \resultcell{0}{0}
            & \resultcell{50}{8}
            & \resultcell{0}{0}
            & \resultcell{0}{0}
            & \resultcell{44}{8}
            & \resultcellbold{60}{7} \\

            \midrule


            \multirow{5}{*}{AntMaze}
            & navigate
            & large
            & \resultcell{24}{2}
            & \resultcell{16}{5}
            & \resultcell{34}{4}
            & \resultcell{75}{6}
            & \resultcell{83}{4}
            & \resultcellbold{91}{2}
            & \resultcellbold{91}{1}
            & \resultcellbold{91}{1} \\

            & navigate
            & giant
            & \resultcell{0}{0}
            & \resultcell{0}{0}
            & \resultcell{0}{0}
            & \resultcell{14}{3}
            & \resultcell{16}{3}
            & \resultcell{65}{5}
            & \resultcell{70}{2}
            & \resultcellbold{78}{2} \\

            \cmidrule(lr){2-11}

            & stitch
            & large
            & \resultcell{3}{3}
            & \resultcell{18}{2}
            & \resultcell{7}{2}
            & \resultcell{18}{2}
            & \resultcell{11}{2}
            & \resultcell{67}{5}
            & \resultcell{79}{3}
            & \resultcellbold{86}{2} \\

            & stitch
            & giant
            & \resultcell{0}{0}
            & \resultcell{0}{0}
            & \resultcell{0}{0}
            & \resultcell{0}{0}
            & \resultcell{0}{0}
            & \resultcell{2}{2}
            & \resultcell{29}{5}
            & \resultcellbold{38}{3} \\

            \cmidrule(lr){2-11}

            & explore
            & large
            & \resultcell{0}{0}
            & \resultcell{10}{3}
            & \resultcell{0}{0}
            & \resultcell{0}{0}
            & \resultcell{0}{0}
            & \resultcell{4}{5}
            & \resultcellbold{62}{12}
            & \resultcellbold{72}{13} \\

            \midrule


            \multirow{4}{*}{HumanoidMaze}
            & navigate
            & large
            & \resultcell{1}{0}
            & \resultcell{2}{1}
            & \resultcell{2}{1}
            & \resultcell{5}{1}
            & \resultcell{24}{4}
            & \resultcell{49}{4}
            & \resultcellbold{82}{2}
            & \resultcell{73}{3} \\

            & navigate
            & giant
            & \resultcell{0}{0}
            & \resultcell{0}{0}
            & \resultcell{0}{0}
            & \resultcell{1}{0}
            & \resultcell{3}{2}
            & \resultcell{12}{4}
            & \resultcellbold{91}{1}
            & \resultcellbold{89}{2} \\

            \cmidrule(lr){2-11}

            & stitch
            & large
            & \resultcell{6}{3}
            & \resultcell{1}{1}
            & \resultcell{0}{0}
            & \resultcell{3}{1}
            & \resultcell{4}{1}
            & \resultcell{28}{3}
            & \resultcell{43}{3}
            & \resultcellbold{51}{4} \\

            & stitch
            & giant
            & \resultcell{0}{0}
            & \resultcell{0}{0}
            & \resultcell{0}{0}
            & \resultcell{0}{0}
            & \resultcell{0}{0}
            & \resultcell{3}{2}
            & \resultcellbold{61}{3}
            & \resultcellbold{63}{3} \\

            \midrule


            \multirow{4}{*}{Visual-AntMaze}
            & navigate
            & large
            & \resultcell{4}{0}
            & \resultcell{5}{1}
            & \resultcell{4}{1}
            & \resultcell{0}{0}
            & \resultcellbold{84}{1}
            & \resultcell{53}{9}
            & \resultcell{63}{4}
            & \resultcell{75}{6} \\

            & navigate
            & giant
            & \resultcell{0}{0}
            & \resultcell{1}{1}
            & \resultcell{0}{0}
            & \resultcell{0}{0}
            & \resultcellbold{47}{2}
            & \resultcell{6}{4}
            & \resultcell{8}{3}
            & \resultcell{21}{5} \\

            \cmidrule(lr){2-11}

            & stitch
            & large
            & \resultcell{24}{3}
            & \resultcell{1}{1}
            & \resultcell{0}{0}
            & \resultcell{1}{1}
            & \resultcell{11}{3}
            & \resultcell{28}{2}
            & \resultcell{21}{2}
            & \resultcellbold{45}{8} \\

            & stitch
            & giant
            & \resultcell{0}{0}
            & \resultcell{0}{0}
            & \resultcell{0}{0}
            & \resultcell{0}{0}
            & \resultcell{0}{0}
            & \resultcell{0}{0}
            & \resultcell{2}{2}
            & \resultcellbold{7}{3} \\

            \midrule


            \multirow{2}{*}{Cube}
            & play
            & single
            & \resultcell{6}{2}
            & \resultcell{53}{4}
            & \resultcellbold{68}{6}
            & \resultcell{5}{1}
            & \resultcell{19}{2}
            & \resultcell{15}{3}
            & \resultcell{12}{2}
            & \resultcell{15}{1} \\

            & play
            & double
            & \resultcell{1}{1}
            & \resultcell{36}{3}
            & \resultcellbold{40}{5}
            & \resultcell{1}{0}
            & \resultcell{10}{2}
            & \resultcell{6}{2}
            & \resultcell{3}{1}
            & \resultcell{3}{1} \\

            \midrule


            Scene
            & play
            & --
            & \resultcell{5}{1}
            & \resultcell{42}{4}
            & \resultcellbold{51}{4}
            & \resultcell{5}{1}
            & \resultcell{19}{2}
            & \resultcell{38}{3}
            & \resultcell{22}{5}
            & \resultcell{26}{6} \\

            \midrule


            \multirow{2}{*}{Visual-Cube}
            & noisy
            & single
            & \resultcell{14}{3}
            & \resultcell{75}{3}
            & \resultcell{48}{3}
            & \resultcell{10}{5}
            & \resultcell{39}{30}
            & \resultcell{99}{0}
            & \resultcell{99}{0}
            & \resultcellbold{100}{0} \\

            & noisy
            & double
            & \resultcell{5}{1}
            & \resultcell{17}{4}
            & \resultcell{22}{2}
            & \resultcell{6}{2}
            & \resultcell{6}{3}
            & \resultcell{59}{3}
            & \resultcell{65}{2}
            & \resultcellbold{75}{1} \\

            \midrule


            Visual-Scene
            & noisy
            & --
            & \resultcell{13}{2}
            & \resultcell{23}{2}
            & \resultcell{12}{4}
            & \resultcell{2}{0}
            & \resultcell{15}{2}
            & \resultcell{50}{1}
            & \resultcell{54}{2}
            & \resultcellbold{61}{1} \\

            \bottomrule
        \end{tabular}

    \end{threeparttable}
\end{table*}

\paragraph{Performance on OGBench.}

Table~\ref{tab:all_baselines_ota_geo} reports success rates across all tasks considered. Aggregated over tasks, GITA outperforms every baseline under a two-sided paired Wilcoxon signed-rank test with Holm-Bonferroni correction for multiple comparisons (all adjusted $p<0.05$; see Appendix~\ref{app:stat_tests}). The locomotion results trace a clear progression from \hiql{} to \ota{} to \gita{}: abstracting the value function at a fixed scale improves over no abstraction, and conditioning on the scale improves further still. The second gap is the one that supports our central claim, that no single temporal abstraction suits all state-goal distances. The gap remains substantial in the \texttt{giant} layouts, where the wider range of state-goal distances within a single environment places strong pressure on methods committed to a single scale. \gita{} also improves on most \texttt{large} layouts, where a fixed scale is already miscalibrated for part of the distance range. \looseness=-1

The exception is state-based manipulation. On Cube and Scene, the non-hierarchical \gciql{} outperforms \gita{} by a wide margin, but it also outperforms \hiql{} and \ota{}, so the gap reflects a broader difficulty of hierarchical methods, rather than of multi-scale abstraction specifically. OGBench attributes this to the \texttt{play} datasets, whose open-loop, non-Markovian trajectories with temporally correlated noise can make policy and subgoal prediction ambiguous. \gita{} still improves over \ota{} on every pixel-based manipulation task, where the \texttt{noisy} datasets are Markovian and provide broader state coverage, yielding more consistent supervision for hierarchical policy learning. \looseness=-1

\begin{table*}[t]
    \centering
    \begin{threeparttable}

        \caption{\textbf{Every variant degrades sharply somewhere; GITA does not.}
        Success rate (\%) on the maze tasks, mean $\pm$ standard deviation over 8 seeds.}
        \label{tab:ablations_best_discount}

        \footnotesize
        \setlength{\tabcolsep}{2.0pt}
        \renewcommand{\arraystretch}{1.18}

        \begin{tabular}{
            @{}
            l
            l
            l
            >{\centering\arraybackslash}p{1.85cm}
            >{\centering\arraybackslash}p{2.00cm}
            >{\centering\arraybackslash}p{2.20cm}
            >{\centering\arraybackslash}p{1.95cm}
            c
            @{}
        }
            \toprule

            \textbf{Environment}
            & \textbf{Type}
            & \textbf{Size}
            & \textbf{Unconditioned Value}
            & \textbf{Highest Advantage Factor}
            & \textbf{Mean Advantage Before Exponentiation}
            & \textbf{GITA with Geometric $\boldsymbol{\factor}$}
            & \textbf{\gita{}} \\

            \midrule


            \multirow{4}{*}{PointMaze}
            & navigate
            & large
            & \resultcell{69}{10}
            & \resultcell{92}{4}
            & \resultcell{94}{3}
            & \resultcell{95}{2}
            & \resultcell{92}{6} \\

            & navigate
            & giant
            & \resultcell{40}{13}
            & \resultcell{77}{6}
            & \resultcell{84}{5}
            & \resultcell{81}{5}
            & \resultcell{82}{5} \\

            \cmidrule(lr){2-8}

            & stitch
            & large
            & \resultcell{37}{16}
            & \resultcell{67}{16}
            & \resultcell{80}{9}
            & \resultcell{65}{15}
            & \resultcell{73}{17} \\

            & stitch
            & giant
            & \resultcell{58}{9}
            & \resultcell{54}{15}
            & \resultcell{55}{16}
            & \resultcell{45}{10}
            & \resultcell{56}{13} \\

            \midrule


            \multirow{5}{*}{AntMaze}
            & navigate
            & large
            & \resultcell{89}{2}
            & \resultcell{92}{3}
            & \resultcell{91}{2}
            & \resultcell{90}{2}
            & \resultcell{91}{2} \\

            & navigate
            & giant
            & \resultcell{62}{4}
            & \resultcell{78}{4}
            & \resultcell{77}{3}
            & \resultcell{80}{4}
            & \resultcell{79}{3} \\

            \cmidrule(lr){2-8}

            & stitch
            & large
            & \resultcell{74}{4}
            & \resultcell{82}{4}
            & \resultcell{85}{2}
            & \resultcell{85}{3}
            & \resultcell{86}{3} \\

            & stitch
            & giant
            & \resultcell{38}{6}
            & \resultcell{29}{5}
            & \resultcell{39}{7}
            & \resultcell{25}{6}
            & \resultcell{42}{6} \\

            \cmidrule(lr){2-8}

            & explore
            & large
            & \resultcell{66}{14}
            & \resultcell{65}{11}
            & \resultcell{73}{13}
            & \resultcell{67}{23}
            & \resultcell{87}{9} \\

            \midrule


            \multirow{4}{*}{HumanoidMaze}
            & navigate
            & large
            & \resultcell{75}{7}
            & \resultcell{74}{4}
            & \resultcell{72}{5}
            & \resultcell{74}{5}
            & \resultcell{72}{5} \\

            & navigate
            & giant
            & \resultcell{89}{2}
            & \resultcell{87}{4}
            & \resultcell{89}{4}
            & \resultcell{90}{3}
            & \resultcell{90}{4} \\

            \cmidrule(lr){2-8}

            & stitch
            & large
            & \resultcell{36}{12}
            & \resultcell{45}{6}
            & \resultcell{52}{6}
            & \resultcell{52}{4}
            & \resultcell{49}{2} \\

            & stitch
            & giant
            & \resultcell{69}{4}
            & \resultcell{62}{7}
            & \resultcell{67}{5}
            & \resultcell{65}{5}
            & \resultcell{68}{3} \\

            \bottomrule
        \end{tabular}

    \end{threeparttable}
\end{table*}

\subsection{Ablation Studies}
\label{sec:ablations}

Table~\ref{tab:ablations_best_discount} compares four variants of GITA, each isolating one component: whether the value function is conditioned on $\factor$, how scale-specific advantages are combined, and how the scales are chosen. \looseness=-1

\textbf{Unconditioned Value} tests whether exposure to multiple temporal scales during value learning suffices on its own, or whether the value function must explicitly represent the abstraction scale. To test this, we remove $\factor$ from the input to the high-level value function while still sampling $\factor \sim\operatorname{Geom}(1-\alpha)$ to construct the abstract transitions in \cref{eq:geo_ota_value}, so the network learns one value function averaged across scales. Only one advantage is then defined per tuple, so the aggregation in \cref{eq:geo_multiscale_weight} reduces to standard AWR; this variant therefore removes both components at once and serves as the cumulative comparison against GITA. \looseness=-1

\textbf{Highest Advantage Factor} tests whether policy supervision is better treated as a hard selection problem, with one scale chosen per sample, than as an aggregation across scales. We test this by retaining the abstraction-conditioned value function but weighting each tuple using only the scale with the largest advantage, $W=\exp(\beta_h \max_{\factor\in\mathcal{K}} A^h_{\factor})$, discarding the rest. \looseness=-1

\textbf{Mean Advantage Before Exponentiation} tests whether the implicit selection induced by exponentiating before averaging matters, or whether exposing the policy to multiple scales suffices. We isolate this effect by keeping all scale-specific advantages but reversing the order, replacing \cref{eq:geo_multiscale_weight} with $\exp(\beta_h \frac{1}{|\mathcal{K}|}\sum_{\factor} A^h)$. Because exponential is convex, this removes the emphasis on large positive scale-specific advantages and instead averages all advantages uniformly before exponentiation. \looseness=-1

\textbf{GITA with Geometric $\boldsymbol{\factor}$} tests whether the high-level policy needs a pre-specified candidate set of abstraction factors. We replace the fixed $\mathcal{K}$ with $|\mathcal{K}|$ factors drawn independently from $\operatorname{Geom}(1-\zeta)$ at every policy update. This removes the spacing of $\mathcal{K}$ as a design choice and ensures the policy is supervised only at scales supported during value learning, though the candidate set is now noisier at each update. \looseness=-1

We observe the largest drop in performance in \emph{Unconditioned Value}. Exposure to multiple temporal scales during value learning is therefore not sufficient on its own; the scale must be explicitly represented in the value function. \gita{} also significantly outperforms \emph{Highest Advantage Factor} ($p<0.05$, Holm-Bonferroni-corrected Wilcoxon signed-rank test; see \Cref{tab:ablation_stat_tests}), so committing to a single scale per sample loses information that soft aggregation retains, even when that scale is the strongest one available. \emph{Mean Advantage Before Exponentiation} and \emph{GITA with Geometric $\factor$} are not significantly different overall, but each collapses on one of AntMaze \texttt{large-explore} and \texttt{giant-stitch} respectively, the two settings with the widest spread of state-goal distances, where \gita{} does not. Neither the exponentiation order nor the fixed candidate set is thus strictly necessary, but together they are what make the method reliable across regimes. \looseness=-1

\section{Conclusion}
In this paper, we introduced Generalized Implicit Temporal Abstraction (GITA), an offline goal-conditioned reinforcement learning (GCRL) method that addresses the trade-off between long-range value propagation and local value discrimination. We showed analytically that the advantage used for subgoal selection factors into a reachability term and a resolution term pulling in opposite directions as the abstraction factor grows, so the best factor increases with state-goal distance, and no fixed choice serves every pair. GITA resolves this by learning a shared value function across temporal scales and training a single policy from their aggregated advantages, implicitly emphasizing the most informative scales for each state-goal pair. Our empirical results establish GITA as a robust and flexible framework for hierarchical offline RL, providing a foundation for future methods that leverage multiple temporal scales to tackle long-horizon tasks.

Directions for extending GITA include: (1) GITA, like other hierarchical methods, is less effective on Cube and Scene, which are short-horizon and compositional rather than long-horizon; incorporating compositionality into the notion of temporal abstraction is a natural next step. (2) In GITA, as in many other approaches, the notion of distance depends on the policy that generated the dataset, and correcting for discrepancies introduced by the off-policy nature of the data is a promising direction. (3) Finally, the representations GITA learns are only implicitly shaped by the value loss; explicitly learning representations that capture an intrinsic notion of distance \citep[e.g.,][]{shehmar2026alps} would likely benefit the framework. \looseness=-1

\subsection*{Acknowledgments}

This research was supported in part by the Natural Sciences and Engineering Research Council of Canada (NSERC), the Canada CIFAR AI Chair Program, and Alberta Innovates. Computational resources were provided in part by the Digital Research Alliance of Canada. This work was also partially funded by CNPq, CAPES, FAPEMIG, IAIA-INCT on AI, INCTTildIAR, and the Brazilian Center on Algorithm Transparency and AI Safety.

\bibliography{iclr2027_conference}
\bibliographystyle{iclr2027_conference}

\clearpage
\appendix

\section{Related Work}
\label{sec:related_work}

\paragraph{Offline goal-conditioned value learning.}
Offline goal-conditioned reinforcement learning (GCRL) aims to learn policies for reaching arbitrary goals from fixed datasets. Existing approaches differ in how they estimate long-range goal-conditioned values: contrastive RL learns state--goal representations through contrastive objectives \citep{eysenbach2022contrastive}, whereas quasimetric RL exploits the structure of temporal distances \citep{wang2023optimal}. Building on the latter, \citet{zheng2026scaling} fit quasimetric distances with multistep Monte Carlo targets to improve long-horizon learning. \citet{jeon2026recursive} instead organize value learning recursively over a balanced decomposition of observed trajectory segments, shortening the chain of bootstrapped dependencies before propagating values across trajectories. These methods address long-range estimation through different objectives.

\paragraph{Hierarchical goal-conditioned reinforcement learning.}
Hierarchical RL addresses long-horizon decision-making by decomposing complex tasks into simpler subtasks. In offline GCRL, \citet{park2023hiql} introduced a two-level policy that proposes intermediate subgoals and executes primitive actions toward them, training both levels from goal-conditioned values using advantage-weighted regression \citep{peng2019advantage}. More recent methods modify the structure or execution of this hierarchy. \citet{choi2026chain} generate sequences of latent subgoals autoregressively before predicting an action, while \citet{ke2026adaptive} recursively refine distant goals until a learned reachability criterion indicates that the current target can be executed locally. \citet{jawaid2026offline} combines high-level subgoal planning with low-level action chunks and temporally extended value backups. \citet{wibault2026abstraction} study relativized options and level-specific representations to reuse experience across similar state--goal contexts. Rather than changing subgoal generation, hierarchy depth, or low-level execution, we retain the two-level policy structure of HIQL and study how temporal abstraction in the high-level value function affects subgoal evaluation.

\paragraph{Temporal abstraction at multiple scales.}
Temporal abstraction has long been formalized through temporally extended actions \citep{sutton1999between}. \citet{ahn2026option} introduced temporally abstracted value learning for offline GCRL, treating a fixed stride of $k$ primitive steps as one abstract transition. This improves long-range value propagation, but a fixed $k$ must trade off sensitivity to distant goals against the ability to discriminate nearby states. Multiple temporal horizons have also been studied in other components of offline RL: \citet{chung2026offline} learn predictive models that can sample future states at arbitrary horizons and use a controlled horizon distribution for model-based value learning, while \citet{adaikkappan2026multiscale} use multiscale predictive supervision to align state and goal representations. Outside offline RL, \citet{sherstan2020gamma} introduced $\Gamma$-nets, which condition a single value estimator on the discount factor $\gamma$, enabling value predictions to generalize across multiple timescales. In contrast, GITA represents multiple temporal abstraction scales directly within a single $k$-conditioned goal-value function, where $k$ controls the number of primitive steps represented by each abstract transition.

\section{Proofs for the Fixed-Scale Resolution--Horizon Trade-off}
\label{app:fixed_scale_tradeoff_derivation}

This section provides the proofs for Proposition~\ref{proposition1} and its corollaries. Our analysis is an idealized shortest-path model and is not intended to exactly characterize dataset-induced transitions. Let \(d=d^\star(s,g)\) denote the minimum number of primitive transitions from state \(s\) to goal \(g\). We follow the temporally abstracted update convention of OTA~\citep{ahn2026option}, which evaluates the sparse reward at the abstract successor \(s_{t+k}\), whereas HIQL indexes the one-step reward at \(s_t\). This results in a one-step reward-indexing shift at \(k=1\), which does not affect the maximizing abstraction factor in the analysis below. Following the theoretical formulation of OTA~\citep{ahn2026option}, the primitive-step value is
\begin{equation}
V^\star(d)
=
-\sum_{j=0}^{d-1}\gamma^j
=
-\frac{1-\gamma^d}{1-\gamma}.
\label{eq:app_primitive_value}
\end{equation}

With abstraction factor \(k\), the corresponding abstract horizon is $H_k(d)=\left\lceil\frac{d}{k}\right\rceil$, which gives
\begin{equation}
V_k^\star(d)
=
-\frac{1-\gamma^{\left\lceil\frac{d}{k}\right\rceil}}{1-\gamma}.
\end{equation}

Relaxing \(\lceil d/k\rceil\) to \(d/k\) so that it varies smoothly with $\factor$, yields
\begin{equation}
\widetilde V_k(s,g)
=
-\frac{1-\gamma^{d^\star(s,g)/k}}{1-\gamma},
\qquad
k>0,
\quad
0<\gamma<1.
\label{eq:app_tradeoff_value}
\end{equation}

Consider a candidate subgoal \(s'\) that makes \(\Delta\) primitive steps of progress toward the goal, such that
\begin{equation}
d^\star(s,g)=d,
\qquad
d^\star(s',g)=d-\Delta,
\qquad
d>\Delta>0.
\label{eq:app_tradeoff_progress}
\end{equation}

\subsection{Proof of Proposition~\ref{proposition1}}

\begin{proof}
The high-level advantage of the candidate subgoal \(s'\) relative to the current state \(s\) is
\begin{equation}
\widetilde A_k(s,s',g)
=
\widetilde V_k(s',g)
-
\widetilde V_k(s,g).
\end{equation}
Applying \Cref{eq:app_tradeoff_value} to both $s$ and $s'$, with distance given by \Cref{eq:app_tradeoff_progress}, we obtain
\begin{align}
\widetilde A_k(s,s',g)
&=
-\frac{1-\gamma^{(d-\Delta)/k}}{1-\gamma}
+
\frac{1-\gamma^{d/k}}{1-\gamma}
\nonumber\\
&=
\frac{\gamma^{(d-\Delta)/k}-\gamma^{d/k}}{1-\gamma}
\nonumber\\
&=
\gamma^{(d-\Delta)/k}
\frac{1-\gamma^{\Delta/k}}{1-\gamma}.
\label{eq:app_advantage_margin}
\end{align}
Therefore,
\begin{equation}
\widetilde A_k(s,s',g)
=
\underbrace{\gamma^{(d-\Delta)/k}}_{\text{reachability}}
\;
\underbrace{\frac{1-\gamma^{\Delta/k}}{1-\gamma}}_{\text{resolution}}
\end{equation}
\end{proof}
The \emph{reachability factor} measures whether the residual distance $d-\Delta$ falls within a range where discounting carries signal, while the \emph{resolution factor} measures how sharply the value function separates $s$ from a subgoal $\Delta$ primitive steps closer.

\subsection{Proof of the Vanishing-Advantage Corollary}

We prove the two limits in \Cref{eq:fixed_scale_extreme_limits} separately. Throughout, let $\lambda := -\ln\gamma > 0$, so that \(\gamma^{x} = e^{-\lambda x}\) for any exponent \(x\).

\paragraph{Limit as \(k\to0^+\).}
\begin{proof}
Since \(d-\Delta>0\) is fixed, $(d-\Delta)/k\to\infty$ as $k\to0^+$, so the reachability factor vanishes:
\begin{equation}
\lim_{k\to0^+}
\gamma^{(d-\Delta)/k}
=
\lim_{k\to0^+}
\exp\left(
-\lambda\frac{d-\Delta}{k}
\right)
=
0.
\label{eq:app_reach_zero}
\end{equation}
Meanwhile the resolution factor stays bounded for all $k$, since $0<\gamma ^{\Delta/k}<1$ implies
\begin{equation}
0
<
\frac{1-\gamma^{\Delta/k}}{1-\gamma}
<
\frac{1}{1-\gamma}.
\label{eq:app_resol_bound}
\end{equation}
Combining Equation (\ref{eq:app_reach_zero}) and (\ref{eq:app_resol_bound}) in \Cref{eq:app_advantage_margin} gives
\begin{equation}
0
<
\widetilde A_k(s,s',g)
<
\frac{1}{1-\gamma}
\exp\left(
-\lambda\frac{d-\Delta}{k}
\right)
\xrightarrow[k\to0^+]{}0
,
\end{equation}
so by the squeeze theorem,
\begin{equation}
\lim_{k\to0^+}
\widetilde A_k(s,s',g)
=
0.
\label{eq:app_fine_scale_limit}
\end{equation}
\end{proof}
Although \(k\) is discrete in the algorithm, this limit concerns its continuous relaxation. Equivalently, the same vanishing regime is obtained for any fixed admissible \(k\) as \(d\to\infty\), since it is governed by the ratio \((d-\Delta)/k\).

\paragraph{Limit as \(k\to\infty\).}
\begin{proof}

Here the reachability factor tends to $1$ while the resolution factor's numerator $1-\gamma^{\Delta/k}$ tends to $0$. Using \(e^{-x}=1-x+\mathcal{O}(x^2)\) as \(x\to0\) with \(x=\lambda\Delta/k\) and \(x=\lambda(d-\Delta)/k\) respectively,
\begin{equation}
  1-\gamma^{\Delta/k}
  =
  \frac{\lambda\Delta}{k}
  +
  \mathcal{O}(k^{-2}),
  \qquad
  \gamma^{(d-\Delta)/k}
  =
  1+\mathcal{O}(k^{-1}).
  \label{eq:app_first_order_expansions}
\end{equation}
Substituting \Cref{eq:app_first_order_expansions} into
\Cref{eq:app_advantage_margin}, all cross terms are \(\mathcal{O}(k^{-2})\)
or smaller, leaving
\begin{equation}
  \widetilde A_k(s,s',g)
  =
  \frac{\lambda\Delta}{(1-\gamma)k}
  +
  \mathcal{O}(k^{-2}).
  \label{eq:app_coarse_scale_expansion}
\end{equation}
Since the leading term is \(\Theta(1/k)\),
\begin{equation}
\lim_{k\to\infty}
\widetilde A_k(s,s',g)
=
0
\label{eq:app_coarse_scale_limit}
\end{equation}
\end{proof}
Together, \Cref{eq:app_fine_scale_limit,eq:app_coarse_scale_limit} prove the vanishing-advantage corollary. Therefore, a coarser abstraction allows the advantage to better discriminate distant states when considering distant goals, because the \emph{reachability factor} increases toward $1$ as $\factor$ grows. However, the same coarser abstraction blurs the distinction between nearby states, since the \emph{resolution factor} decreases toward $0$ as $k$ grows. No $\factor$ simultaneously maximizes both. \looseness=-1

\subsection{Proof of Corollary~\ref{cor:optimal}}
\begin{proof}
From \Cref{eq:app_advantage_margin},
\begin{equation}
\widetilde A_k(d,\Delta)
=
\frac{\gamma^{(d-\Delta)/k}-\gamma^{d/k}}{1-\gamma}.
\label{eq:app_advantage_difference_form}
\end{equation}
Differentiating with respect to \(k\) gives
\begin{equation}
\frac{\partial\widetilde A_k(d,\Delta)}{\partial k}
=
\frac{-\log\gamma}{(1-\gamma)k^2}
\left[
(d-\Delta)\gamma^{(d-\Delta)/k}
-
d\gamma^{d/k}
\right].
\label{eq:app_advantage_derivative}
\end{equation}
Since $-\log\gamma>0$, $1-\gamma>0$, and $k^2>0$, the sign of the derivative is governed entirely by the terms in brackets, and a stationary point satisfies
\begin{equation}
(d-\Delta)\gamma^{(d-\Delta)/k}
=
d\gamma^{d/k}.
\end{equation}
Dividing both sides by \(\gamma^{(d-\Delta)/k}>0\) and using $\gamma^{d/k}=\gamma^{(d-\Delta)/k}\gamma^{\Delta/k}$,
\begin{equation}
d-\Delta
=
d\gamma^{\Delta/k}
\qquad \Longrightarrow \qquad \gamma^{\Delta/k}
=
\frac{d-\Delta}{d}.
\label{eq:app_stationary_condition}
\end{equation}
Taking logarithms and solving for \(k\) gives the exact continuous optimum
\begin{equation}
k^\star_{\mathrm{opt}}(d,\Delta)
=
\frac{-\Delta\log\gamma}
{\log\left(\frac{d}{d-\Delta}\right)}.
\label{eq:app_optimal_abstraction_factor}
\end{equation}

Since \(\gamma^{\Delta/k}\) increases strictly and continuously from \(0\) to \(1\) as \(k\) ranges over \((0, \infty)\), and \((d-\Delta)/d\in(0,1)\), \Cref{eq:app_stationary_condition} has a unique solution. From \Cref{eq:app_advantage_derivative}, the derivative is positive before this solution and negative after it. Therefore, \(k^\star_{\mathrm{opt}}(d,\Delta)\) is the unique global maximizer of the relaxed advantage.
\end{proof}
\paragraph{Dependence on state-goal distance.}

For fixed \(\Delta\) and \(\gamma\), write $k^\star_{\mathrm{opt}}(d,\Delta) = C/L(d)$ with $C=-\Delta\log\gamma>0$ and $L(d)
=
\log\left(
\frac{d}{d-\Delta}
\right).$
Since
\begin{equation}
L'(d)
=
\frac{1}{d}
-
\frac{1}{d-\Delta}
=
-\frac{\Delta}{d(d-\Delta)}
<
0,
\end{equation}
the quotient rule gives
\begin{equation}
\frac{\partial k^\star_{\mathrm{opt}}}{\partial d}
=
\frac{(-\Delta\log\gamma)\Delta}
{d(d-\Delta)
\left[
\log\left(\frac{d}{d-\Delta}\right)
\right]^2}
>
0.
\label{eq:app_optimal_factor_monotonic_distance}
\end{equation}
Thus, the advantage-maximizing abstraction factor increases with state-goal distance. This monotonicity result holds for fixed \(\Delta\) and \(\gamma\).

\paragraph{Local-progress approximation.}

For the regime considered in Corollary~\ref{cor:optimal}, let $\epsilon
=
\frac{\Delta}{d}
\ll1.$
Expanding the denominator of \Cref{eq:app_optimal_abstraction_factor},
\begin{align}
\log\left(
\frac{d}{d-\Delta}
\right)
&=
-\log(1-\epsilon)
\nonumber\\
&=
\epsilon
+
\frac{\epsilon^2}{2}
+
\mathcal{O}(\epsilon^3).
\label{eq:app_log_expansion}
\end{align}
Substituting \(\Delta=d\epsilon\) into \Cref{eq:app_optimal_abstraction_factor} and dividing through by $\epsilon$,
\begin{equation}
  k^\star_{\mathrm{opt}}(d,\Delta)
  =
  \frac{-d\epsilon\log\gamma}
  {\epsilon\left[1+\epsilon/2+\mathcal{O}(\epsilon^2)\right]}
  =
  \frac{-d\log\gamma}
  {1+\epsilon/2+\mathcal{O}(\epsilon^2)}.
  \label{eq:app_optimal_factor_intermediate}
\end{equation}
Applying the reciprocal expansion \(1/(1+x)=1-x+\mathcal{O}(x^2)\) as
\(x\to0\), with \(x=\epsilon/2\), gives
\begin{equation}
  k^\star_{\mathrm{opt}}(d,\Delta)
  =
  -d\log\gamma\left[1-\frac{\epsilon}{2}+\mathcal{O}(\epsilon^2)\right].
  \label{eq:app_optimal_factor_expansion}
\end{equation}
Therefore,
\begin{equation}
k^\star_{\mathrm{opt}}(d,\Delta)
\approx
-d\log\gamma,
\qquad
\Delta\ll d
\label{eq:app_optimal_factor_local}
\end{equation}
which proves \Cref{eq:fixed_scale_optimal_factor_local}.

\paragraph{Effective discount horizon.}

Dividing \(d-\Delta=d(1-\epsilon)\) by \Cref{eq:app_optimal_factor_expansion} and re-expanding to first order in \(\epsilon\) gives
\begin{equation}
\frac{d-\Delta}{k^\star_{\mathrm{opt}}(d,\Delta)}
=
-\frac{1}{\log\gamma}
\left[
1-\frac{\epsilon}{2}
+\mathcal{O}(\epsilon^2)
\right],
\label{eq:app_effective_horizon_expansion}
\end{equation}
so that, for all \(\Delta\ll d\),
\begin{equation}
\frac{d-\Delta}{k^\star_{\mathrm{opt}}(d,\Delta)}
\approx
-\frac{1}{\log\gamma}.
\end{equation}
Since $-1/\log\gamma \sim 1/(1-\gamma)$ as \(\gamma\to1\), combining the two approximations yields
\begin{equation}
\frac{d-\Delta}{k^\star_{\mathrm{opt}}(d,\Delta)}
\approx
-\frac{1}{\log\gamma}
\approx
\frac{1}{1-\gamma},
\qquad
\Delta\ll d,
\quad
\gamma\to1,
\label{eq:app_effective_horizon}
\end{equation}
which proves \Cref{eq:fixed_scale_effective_horizon}.

Together, these results show that no fixed abstraction factor can preserve informative advantages across all state-goal distances: small \(\factor\) loses signal for distant goals, while large \(\factor\) sacrifices local resolution, and the best scale therefore grows with distance.

\section{Fixed-Scale Trade-off Experiment Details}
\label{app:fixed_scale_tradeoff}

This section provides additional details on the experiment used to evaluate how the quality of the high-level advantage varies with the abstraction factor and the state-goal distance.

\paragraph{Geodesic distance.}
For each environment, we sample $(s,s',g)$ triples using the same sampling procedure used to construct training examples for the high-level policy. We then compute reference distances using the known geometry of the OGBench mazes. Each maze is discretized into a fine grid, where navigable cells form the nodes of a four-connected graph and edges connect adjacent cells without crossing walls. The geodesic distance between two states is defined as the shortest-path distance between their corresponding cells and is computed using Dijkstra's algorithm.

For each triple, we compute $d_{\mathrm{geo}}(s,g)$ and $d_{\mathrm{geo}}(s',g)$ and define the geodesic progress as
\begin{equation}
\Delta d
=
d_{\mathrm{geo}}(s,g)
-
d_{\mathrm{geo}}(s',g).
\label{eq:app_geodesic_progress}
\end{equation}
The triples are grouped into equal-width intervals. Geodesic distance is used only as a proxy for task progress and need not equal expert-policy temporal distance.

\paragraph{Advantage--Progress Correlation.}
For each abstraction factor $k$, we compute the scale-conditioned high-level advantage
\begin{equation}
A_k^h(s,s',g)
=
V_k^h(s',g)-V_k^h(s,g).
\label{eq:app_learned_advantage}
\end{equation}
To measure how well this advantage ranks subgoals according to their progress toward the goal, we compute the Spearman rank correlation between $A_k^h(s,s',g)$ and the corresponding geodesic progress $\Delta d$. We first group training triples $(s,s',g)$ according to their geodesic state-goal distance $d_{\mathrm{geo}}(s,g)$. Each distance bin $b$ corresponds to an interval of state-goal distances, and $\mathcal{I}_b$ denotes the set of triples whose $d_{\mathrm{geo}}(s,g)$ falls within that interval. For each abstraction factor $k$ and distance bin $b$, we define the Advantage-Progress Correlation (APC) as
\begin{equation}
\mathrm{APC}_{k,b}
=
\rho_{\mathrm{S}}
\left(
\left\{A_{k,i}^{h}\right\}_{i\in\mathcal{I}_b},
\left\{\Delta d_i\right\}_{i\in\mathcal{I}_b}
\right).
\label{eq:app_progress_correlation}
\end{equation}
Values close to $1$ indicate that the learned advantage preserves the ordering induced by geodesic progress, values near $0$ indicate little monotonic agreement, and negative values indicate an inverted ordering. We compute APC independently for each training seed and report the mean and one standard deviation across four seeds.

\section{Environments}
\label{app:environments}

This section provides additional details on the environments and offline datasets used in our experiments. We evaluate GITA on locomotion and manipulation tasks from OGBench~\citep{park2025ogbench}. Our locomotion experiments use PointMaze, AntMaze, and HumanoidMaze, which combine maze navigation with increasingly complex control, while our manipulation experiments use Cube and Scene. We consider state-based observations for all five environments and pixel-based variants of AntMaze, Cube, and Scene.

\subsection{OGBench Environments}
\label{app:ogbench_environments}

\paragraph{Maze navigation.}
PointMaze controls a 2-D point mass directly through planar actions, AntMaze controls an 8-DoF quadruped, and HumanoidMaze controls a 21-DoF humanoid. The task in each case is to reach a target location in the maze, with AntMaze and HumanoidMaze additionally requiring the agent to learn the underlying locomotion dynamics. State-based observations contain the full low-dimensional state, including the agent's \(x\)-\(y\) position. In Visual-AntMaze, the low-dimensional observation is replaced by a \(64\times64\times3\) RGB image rendered from a third-person camera, without additional proprioceptive state information.

\paragraph{Maze layouts.}
OGBench provides the \texttt{medium}, \texttt{large}, and \texttt{giant} layouts. The \texttt{medium} and \texttt{large} layouts follow the corresponding D4RL \citep{fu2020d4rl} maze structures, while \texttt{giant} is approximately twice the size of \texttt{large} and contains substantially longer paths. These layouts provide increasingly long state-goal distances and therefore directly stress long-horizon value propagation. The main benchmark focuses primarily on the \texttt{large} and \texttt{giant} layouts.

\begin{figure}[t]
    \centering
    \includegraphics[width=\linewidth]{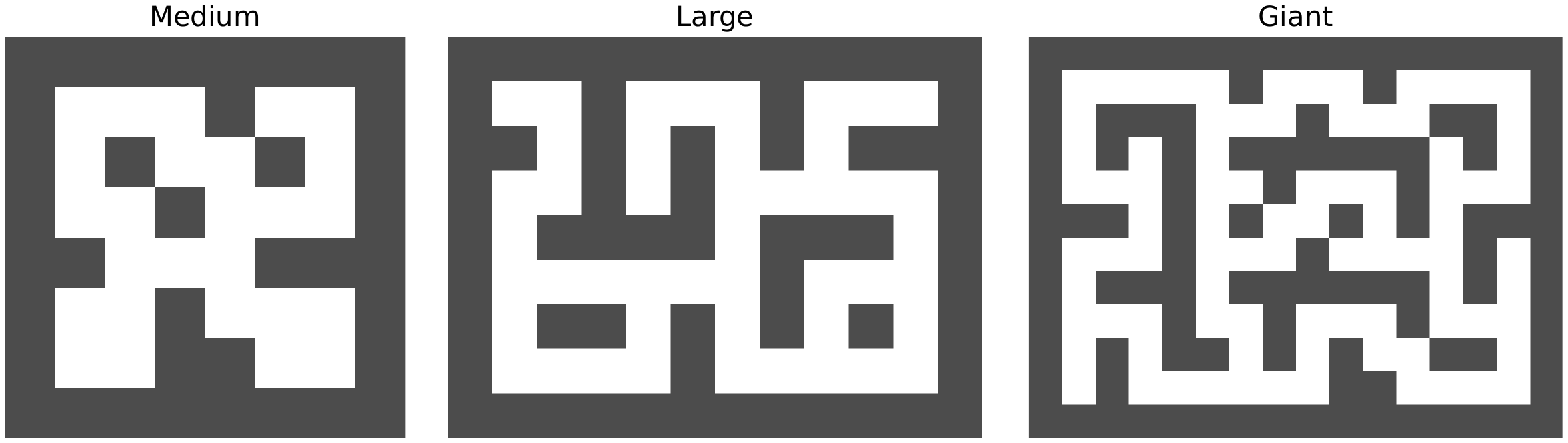}
    \caption{\textbf{Maze layouts.} OGBench provides the \texttt{medium}, \texttt{large}, and \texttt{giant} layouts considered in our experiments, with progressively longer state-goal distances.}
    \label{fig:appendix_maze_layouts}
\end{figure}

\paragraph{Manipulation.}
Cube and Scene use a UR5e robot arm controlled through a 5-dimensional end-effector action space. Cube requires arranging blocks into target configurations and provides variants with different numbers of cubes; our experiments use the \texttt{single} and \texttt{double} variants for state observations. Scene contains a cube, a drawer, a window, and two locking buttons, and its evaluation goals require composing multiple manipulation behaviors into a desired object configuration. Both environments also support \(64\times64\times3\) RGB observations; in the visual variants, the arm is rendered transparently to reduce visual occlusion.

\subsection{Offline Dataset Variants}
\label{app:dataset_variants}

\begin{figure}[t]
    \centering
    \includegraphics[width=\linewidth]{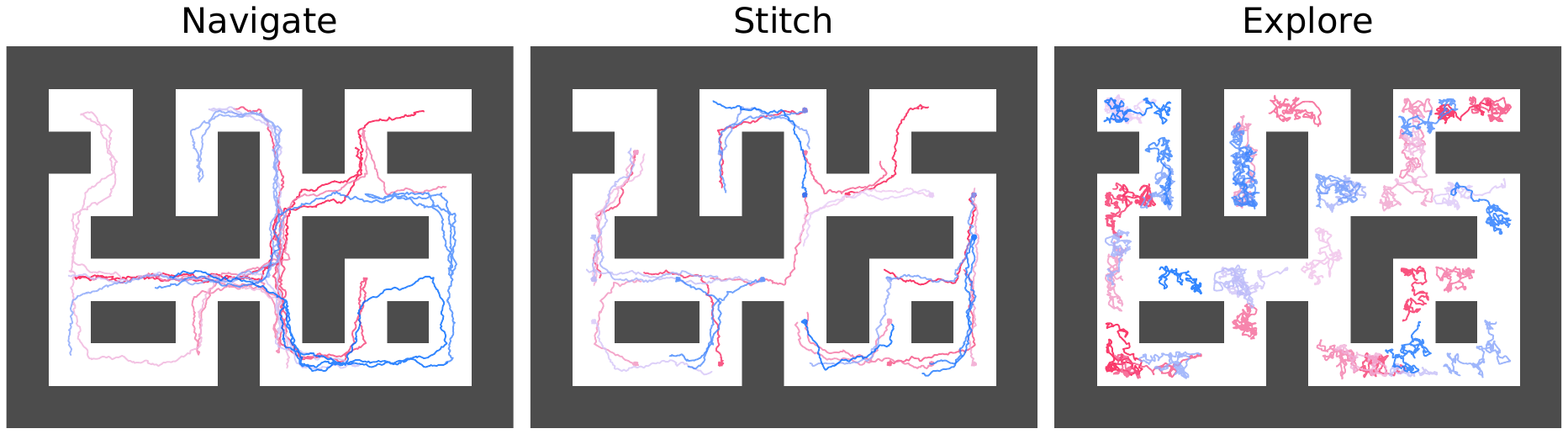}
    \caption{\textbf{OGBench datasets exhibit distinct trajectory structures.} Representative trajectories from the OGBench offline dataset variants in the \texttt{large} maze. The \texttt{navigate} dataset contains long goal-directed trajectories, \texttt{stitch} consists of shorter local trajectory segments that must be combined to solve long-horizon tasks, and \texttt{explore} contains high-coverage but highly suboptimal behavior.}
    \label{fig:appendix_dataset_types}
\end{figure}

The maze environments use three offline dataset variants with different trajectory structure and behavioral quality. The \texttt{navigate} datasets are collected by a noisy expert that repeatedly moves toward randomly sampled goals, producing relatively long and coherent goal-reaching trajectories. The \texttt{stitch} datasets instead consist of short local trajectory segments, so solving evaluation tasks requires combining information across multiple trajectories. For AntMaze, we additionally consider the \texttt{explore} dataset, which contains high-coverage but suboptimal behavior generated from randomly changing movement directions with substantial action noise. State- and pixel-based AntMaze variants share the same underlying trajectories and differ only in their observation modality. For manipulation, OGBench provides \texttt{play} and \texttt{noisy} datasets: \texttt{play} is generated by non-Markovian scripted policies with temporally correlated noise, whereas \texttt{noisy} uses Markovian scripted policies with uncorrelated Gaussian noise and higher state coverage. Our state-based Cube and Scene experiments use \texttt{play}, while Visual-Cube and Visual-Scene use \texttt{noisy}. Figure~\ref{fig:appendix_dataset_types} illustrates the characteristic trajectory structure of the maze dataset variants.

Table~\ref{tab:environment_specs} summarizes the observation and action spaces and evaluation horizons of the environment variants used in our experiments. Table~\ref{tab:dataset_specs} reports the corresponding dataset statistics. Note that OGBench dataset episode lengths need not coincide with the maximum evaluation episode length.

\begin{table*}[t]
    \centering
    \caption{\textbf{Environment specifications.} Specifications for the OGBench domains considered in our experiments.}
    \label{tab:environment_specs}
    \small

    \setlength{\tabcolsep}{5pt}
    \renewcommand{\arraystretch}{1.12}
    \begin{tabular}{lllcc}
        \toprule
        \textbf{Environment} & \textbf{Variant} & \textbf{Observation} & \textbf{Action Dim.} & \textbf{Max. Episode Length} \\
        \midrule

        \multirow{2}{*}{PointMaze}
        & \texttt{large} & state (2) & 2 & 1000 \\
        & \texttt{giant} & state (2) & 2 & 1000 \\

        \midrule

        \multirow{2}{*}{AntMaze}
        & \texttt{large} & state (29) / pixels (\(64\times64\times3\)) & 8 & 1000 \\
        & \texttt{giant} & state (29) / pixels (\(64\times64\times3\)) & 8 & 1000 \\

        \midrule

        \multirow{2}{*}{HumanoidMaze}
        & \texttt{large} & state (69) & 21 & 2000 \\
        & \texttt{giant} & state (69) & 21 & 4000 \\

        \midrule

        \multirow{2}{*}{Cube}
        & \texttt{single} & state (28) / pixels (\(64\times64\times3\)) & 5 & 200 \\
        & \texttt{double} & state (37) / pixels (\(64\times64\times3\)) & 5 & 500 \\

        \midrule

        Scene
        & -- & state (40) / pixels (\(64\times64\times3\)) & 5 & 750 \\

        \bottomrule
    \end{tabular}
\end{table*}

\begin{table*}[t]
    \centering
    \caption{\textbf{Offline dataset specifications.} Specifications of the OGBench offline datasets considered in our experiments. Visual-AntMaze uses the same trajectory datasets and statistics as the corresponding AntMaze configuration.}
    \label{tab:dataset_specs}
    \small
    \setlength{\tabcolsep}{5pt}
    \renewcommand{\arraystretch}{1.12}

    \begin{tabular}{lllccc}
        \toprule
        \textbf{Environment}
        & \textbf{Dataset}
        & \textbf{Variant}
        & \textbf{\# Transitions}
        & \textbf{\# Episodes}
        & \textbf{Data Episode Length} \\
        \midrule

        \multirow{4}{*}{PointMaze}
        & \texttt{navigate} & \texttt{large} & 1M & 1000 & 1000 \\
        & \texttt{navigate} & \texttt{giant} & 1M & 500 & 2000 \\
        \cmidrule(lr){2-6}
        & \texttt{stitch} & \texttt{large} & 1M & 5000 & 200 \\
        & \texttt{stitch} & \texttt{giant} & 1M & 5000 & 200 \\

        \midrule

        \multirow{5}{*}{AntMaze}
        & \texttt{navigate} & \texttt{large} & 1M & 1000 & 1000 \\
        & \texttt{navigate} & \texttt{giant} & 1M & 500 & 2000 \\
        \cmidrule(lr){2-6}
        & \texttt{stitch} & \texttt{large} & 1M & 5000 & 200 \\
        & \texttt{stitch} & \texttt{giant} & 1M & 5000 & 200 \\
        \cmidrule(lr){2-6}
        & \texttt{explore} & \texttt{large} & 5M & 10000 & 500 \\

        \midrule

        \multirow{4}{*}{HumanoidMaze}
        & \texttt{navigate} & \texttt{large} & 2M & 1000 & 2000 \\
        & \texttt{navigate} & \texttt{giant} & 4M & 1000 & 4000 \\
        \cmidrule(lr){2-6}
        & \texttt{stitch} & \texttt{large} & 2M & 5000 & 400 \\
        & \texttt{stitch} & \texttt{giant} & 4M & 10000 & 400 \\

        \midrule

        \multirow{2}{*}{Cube}
        & \texttt{play} & \texttt{single} & 1M & 1000 & 1000 \\
        & \texttt{play} & \texttt{double} & 1M & 1000 & 1000 \\

        \midrule

        Scene
        & \texttt{play} & -- & 1M & 1000 & 1000 \\

        \midrule

        \multirow{2}{*}{Visual-Cube}
        & \texttt{noisy} & \texttt{single} & 1M & 1000 & 1000 \\
        & \texttt{noisy} & \texttt{double} & 1M & 1000 & 1000 \\

        \midrule

        Visual-Scene
        & \texttt{noisy} & -- & 1M & 1000 & 1000 \\

        \bottomrule
    \end{tabular}
\end{table*}

\subsection{Evaluation Protocol}
\label{app:evaluation_protocol}

We follow the OGBench evaluation protocol~\citep{park2025ogbench}, using five predefined state--goal evaluation tasks for each benchmark environment. At each evaluation checkpoint, we run \(50\) episodes for each of the five tasks, yielding \(250\) evaluation episodes per checkpoint and seed. We evaluate state-based experiments over \(8\) random seeds and pixel-based experiments over \(4\) random seeds, and report the mean success rate and standard deviation across seeds.

The results in our main table are averaged over the final three evaluation checkpoints: for state-based tasks, models are trained for \(10^6\) gradient steps and evaluated at \(800\text{K}\), \(900\text{K}\), and \(1\text{M}\) steps. For pixel-based tasks, models are trained for \(500\text{K}\) gradient steps and evaluated at \(300\text{K}\), \(400\text{K}\), and \(500\text{K}\) steps. Thus, each reported per-seed score averages performance over \(3\times5\times50=750\) evaluation episodes before results are aggregated across seeds.

 For the ablation studies, we instead report performance using only the final training checkpoint (\(1\text{M}\) steps for state-based tasks).

\section{Implementation Details}
\label{app:implementation_details}

We implement GITA on top of the HIQL implementation provided by OGBench, which facilitates reproducibility and enables direct comparison with baselines evaluated under the same implementation framework. Unless explicitly stated, we retain the HIQL network architectures, optimization settings, goal-sampling distributions, and hyperparameters. The main architectural change is restricted to the high-level value function, which is additionally conditioned on the temporal abstraction factor \(k\). The high-level and low-level policies retain the HIQL architectures.

\paragraph{Network architecture.}
Following the OGBench implementation of HIQL, the value and policy networks use three hidden layers of size \((512,512,512)\) with Gaussian Error Linear Unit (GELU) activations~\citep{Hendrycks2016Gaussian} and Layer Normalization~\citep{Ba2016Layer}. For state-based tasks, the goal representation network maps the concatenated state-goal input to a \(10\)-dimensional length-normalized representation using the same hidden dimensions. The low-level policy predicts primitive actions conditioned on the current state and the learned subgoal representation, while the high-level policy predicts a \(10\)-dimensional subgoal representation.

For pixel-based tasks, we encode images with the \texttt{impala\_small} encoder, based on the IMPALA architecture~\citep{Espeholt2018IMPALA}, before the corresponding value, policy, and goal-representation networks. This encoder uses three residual stacks with \(16\), \(32\), and \(32\) channels, one residual block per stack, followed by a \(512\)-dimensional MLP. We use image augmentation with probability \(0.5\) for visual manipulation tasks and no augmentation for the visual maze tasks.

The value function consists of two independently parameterized scalar heads, \(V_1\) and \(V_2\), together with corresponding target networks. The minimum of the two next-state target values is used when constructing the conservative bootstrap signal, while policy advantages are computed from the mean of the two online value estimates.

\paragraph{Abstraction-factor conditioning.}
Only the high-level value function receives the abstraction factor as input. We provide \(k\) to the network on a logarithmic scale and compute a \(32\)-dimensional abstraction embedding
\begin{equation}
e_k=f_{\mathrm{abs}}(\log k),
\end{equation}
where \(f_{\mathrm{abs}}\) is a one-layer MLP with GELU activation. We use \(\log k\) because the abstraction factors span multiplicative temporal scales; the logarithm converts multiplicative changes in \(k\) into approximately additive changes in the conditioning input and reduces the dynamic range seen by the embedding network.

We inject this embedding into the hidden layers of the high-level value function using Feature-wise Linear Modulation~\citep[FiLM;][]{perez2018film}. For a hidden representation \(h_\ell\) at layer \(\ell\), the abstraction embedding is projected into scale and shift parameters,
\begin{equation}
\left[\eta_\ell(k),\,\delta_\ell(k)\right]
=
W_\ell^{\mathrm{FiLM}}e_k+b_\ell^{\mathrm{FiLM}},
\end{equation}
and the hidden representation is modulated as
\begin{equation}
h_\ell
\leftarrow
\left(1+\eta_\ell(k)\right)\odot h_\ell+\delta_\ell(k).
\label{eq:app_film}
\end{equation}
FiLM is applied after the activation and LayerNorm in the high-level value network. The FiLM projection layers are initialized to zero.

The state-goal representation is independent of the abstraction factor: \(k\) does not modify the state or goal encoder and is introduced only through FiLM in the high-level value function. Since \(k\) is introduced only through FiLM, the same state-goal representation can be reused across abstraction factors, allowing GITA to support multiple temporal scales with only modest additional training cost. \looseness=-1

\paragraph{Optimization and hyperparameters.}
All networks are optimized with Adam~\citep{kingma2015adam} using a learning rate of \(3\times10^{-4}\). Following the OGBench protocol, state-based models use a minibatch size of \(1024\) and are trained for \(10^6\) gradient steps, while pixel-based models use a minibatch size of \(256\) and are trained for \(5\times10^5\) gradient steps. The target value networks are updated using Polyak averaging with coefficient \(\tau=0.005\), and the expectile parameter is fixed to \(0.7\). \looseness=-1

We retain the HIQL subgoal offset \(m\), independently of the abstraction factor \(k\). We use \(m=25\) for PointMaze and AntMaze, \(m=100\) for HumanoidMaze, and \(m=10\) for Cube and Scene, with the same offsets used for their corresponding visual variants. The subgoal representation dimension is \(10\). \looseness=-1

We preserve the environment-specific HIQL policy-extraction and goal-sampling hyperparameters from OGBench. In particular, the AWR inverse temperature is \(3\) for the standard \texttt{navigate} and \texttt{stitch} datasets and \(10\) for AntMaze \texttt{explore}. For value learning, goals are sampled as the current state, a geometrically sampled future state, or a random dataset state with probabilities \(0.2\), \(0.5\), and \(0.3\), respectively. For high-level actor training, \texttt{navigate} uses trajectory goals, \texttt{stitch} uses an equal mixture of trajectory and random goals, and \texttt{explore} uses random goals.

The discount factor $\gamma$ is the only HIQL hyperparameter that we re-tune for GITA. The OGBench reference implementation already uses environment-dependent discount factors for HIQL, with $\gamma=0.995$ for the longest-horizon maze environments and $\gamma=0.99$ otherwise. In GITA, we re-tune $\gamma$ because it plays an additional role beyond TD learning: we use the same value both as the discount factor in the temporal-difference updates and as the parameter controlling the geometric distribution over abstraction factors. We therefore sweep $\gamma \in \{0.98, 0.99, 0.995\}$ and tune it independently for each task. For simplicity, we chose $\alpha=\gamma$. For the manipulation tasks, we additionally consider $\gamma=0.97$. All remaining HIQL hyperparameters are kept unchanged.

For the candidate abstraction factors, we use values from the Fibonacci sequence, starting from \(3\). This gives \(\mathcal{K}=\{3,5,8,13,21\}\). The range was chosen to cover the abstraction factors used by the fixed-scale OTA baseline: the smallest candidate, \(3\), is below the smallest task-specific abstraction factor selected for OTA, while \(21\) extends beyond the largest fixed factor. For OTA, we use the results reported under the unified-hyperparameter setting, except for the abstraction factor, which is selected separately for each task.

\begin{table}[t]
    \centering
    \caption{\textbf{Hyperparameters used for GITA.} Unless marked as GITA-specific, the values follow the HIQL reference implementation in OGBench.}
    \label{tab:geo_ota_hyperparameters}
    \small
    \setlength{\tabcolsep}{5.5pt}
    \renewcommand{\arraystretch}{1.12}
    \begin{tabular}{ll}
        \toprule
        \textbf{Hyperparameter} & \textbf{Value} \\
        \midrule

        \multicolumn{2}{c}{\textcolor{gray!65}{\textit{-- Optimization and Architecture --}}} \\
        Learning rate & $3\times10^{-4}$ \\
        Optimizer & Adam \\
        Gradient steps & $10^6$ (state), $5\times10^5$ (pixels) \\
        Minibatch size & $1024$ (state), $256$ (pixels) \\
        Actor hidden dimensions & $(512,512,512)$ \\
        Value hidden dimensions & $(512,512,512)$ \\
        Nonlinearity & GELU \\
        Value LayerNorm & True \\
        Visual encoder & \texttt{impala\_small} \\
        Image augmentation probability & $0.5$ (visual manipulation), $0$ (otherwise) \\
        Target smoothing coefficient $\tau$ & $0.005$ \\
        Expectile & $0.7$ \\
        Subgoal representation dimension & $10$ \\
        Subgoal offset $m$ & $25$ (PointMaze, AntMaze), $100$ (HumanoidMaze), $10$ (Cube, Scene) \\

        \multicolumn{2}{c}{\textcolor{gray!65}{\textit{-- Goal Sampling and Policy Extraction --}}} \\
        Value current-goal probability & $0.2$ \\
        Value future-goal probability & $0.5$ (geometric) \\
        Value random-goal probability & $0.3$ \\
        High-level actor goals (\texttt{navigate}) & trajectory: $1.0$ \\
        High-level actor goals (\texttt{stitch}) & trajectory: $0.5$, random: $0.5$ \\
        High-level actor goals (\texttt{explore}) & random: $1.0$ \\
        AWR inverse temperature & $3$ (\texttt{navigate}/\texttt{stitch}), $10$ (\texttt{explore}) \\

        \multicolumn{2}{c}{\textcolor{gray!65}{\textit{-- GITA-Specific Parameters --}}} \\
        Discount factor $\gamma$ & $\{0.97, 0.98,0.99,0.995\}$ (tuned per task) \\
        Geometric sampling parameter $\alpha$ & $\alpha=\gamma$ \\
        Candidate abstraction factors $\mathcal{K}$ & $\{3,5,8,13,21\}$ \\
        Abstraction embedding dimension & $32$ \\

        \bottomrule
    \end{tabular}
\end{table}
\section{Statistical Testing}
\label{app:stat_tests}

We assess differences between GITA and each baseline using a two-sided paired Wilcoxon signed-rank test. Each task constitutes one paired observation, using the mean success rate reported in the corresponding results table; thus, the main comparison uses \(23\) tasks and the ablation comparison uses \(13\) tasks. Because multiple baselines are compared against GITA, we correct the resulting \(p\)-values using the Holm--Bonferroni step-down procedure. Corrections are applied separately within each family of comparisons: the seven baselines in the main results and the four variants in the ablation study. We consider a difference statistically significant when the Holm-adjusted \(p\)-value is below \(0.05\).

\begin{table}[t]
    \centering
    \caption{\textbf{Two-sided paired Wilcoxon signed-rank tests comparing GITA with the baselines in \Cref{tab:all_baselines_ota_geo}.} Each test is performed over the \(23\) tasks reported in the main results table. \(p_{\mathrm{adj}}\) denotes the Holm--Bonferroni-adjusted \(p\)-value across the seven comparisons.}
    \label{tab:main_stat_tests}
    \small
    \setlength{\tabcolsep}{6pt}
    \renewcommand{\arraystretch}{1.12}
    \begin{tabular}{lccc}
        \toprule
        \textbf{Comparison} & \textbf{\(W\)} & \textbf{Raw \(p\)} & \textbf{\(p_{\mathrm{adj}}\)} \\
        \midrule
        GITA vs.\ \gcbc
        & \(0.0\)
        & \(2.69\times10^{-5}\)
        & \(1.88\times10^{-4}\) \\

        GITA vs.\ \gcivl
        & \(14.0\)
        & \(1.62\times10^{-4}\)
        & \(8.08\times10^{-4}\) \\

        GITA vs.\ \gciql
        & \(18.5\)
        & \(2.78\times10^{-4}\)
        & \(8.08\times10^{-4}\) \\

        GITA vs.\ \qrl
        & \(6.5\)
        & \(6.31\times10^{-5}\)
        & \(3.79\times10^{-4}\) \\

        GITA vs.\ \crl
        & \(17.0\)
        & \(2.32\times10^{-4}\)
        & \(8.08\times10^{-4}\) \\

        GITA vs.\ \hiql
        & \(7.0\)
        & \(1.62\times10^{-4}\)
        & \(8.08\times10^{-4}\) \\

        GITA vs.\ \ota
        & \(16.0\)
        & \(5.36\times10^{-4}\)
        & \(8.08\times10^{-4}\) \\
        \bottomrule
    \end{tabular}
\end{table}
All comparisons remain significant after correction, supporting the aggregate improvement of GITA over each baseline in the main evaluation. And GITA significantly improves over removing scale conditioning and hard scale selection, while the two alternative aggregation/sampling variants are not significantly different after correction.

\begin{table}[t]
    \centering
    \caption{\textbf{Two-sided paired Wilcoxon signed-rank tests comparing GITA with the variants in \Cref{tab:ablations_best_discount}.} \(p_{\mathrm{adj}}\) denotes the Holm--Bonferroni-adjusted \(p\)-value across the four comparisons.}
    \label{tab:ablation_stat_tests}
    \small
    \setlength{\tabcolsep}{5pt}
    \renewcommand{\arraystretch}{1.12}
    \begin{tabular}{lccc}
        \toprule
        \textbf{Comparison} & \textbf{\(W\)} & \textbf{Raw \(p\)} & \textbf{\(p_{\mathrm{adj}}\)} \\
        \midrule
        GITA vs.\ Unconditioned Value
        & \(10.0\) & \(0.0100\) & \(0.0300\) \\
        GITA vs.\ Highest Advantage Factor
        & \(5.0\) & \(0.00537\) & \(0.0215\) \\
        GITA vs.\ Mean Advantage Before Exponentiation
        & \(30.5\) & \(0.873\) & \(0.873\) \\
        GITA vs.\ Geometric \(k\)
        & \(21.5\) & \(0.186\) & \(0.371\) \\
        \bottomrule
    \end{tabular}
\end{table}

\section{Baselines}
\label{app:baselines}

This section provides additional details on the baseline methods considered in our experiments.

\subsection{OGBench Baselines}
\label{app:ogbench_baselines}

We compare against the six offline GCRL algorithms benchmarked in OGBench~\citep{park2025ogbench}. Goal-conditioned behavioral cloning~\citep[GCBC;][]{ghosh2019learning} trains a goal-conditioned policy by relabeling transitions with future states from the same trajectory as goals. Goal-conditioned implicit value learning and goal-conditioned implicit Q-learning~\citep[GCIVL and GCIQL;][]{kostrikov2021offline,park2023hiql} are goal-conditioned variants of Implicit Q-Learning (IQL). Both use expectile regression to estimate optimal goal-conditioned value functions. GCIQL learns both a state-value function \(V(s,g)\) and an action-value function \(Q(s,a,g)\), whereas GCIVL is the value-only variant introduced with HIQL and learns only \(V(s,g)\). Quasimetric reinforcement learning~\citep[QRL;][]{wang2023optimal} models goal-reaching distance with a learned quasimetric and exploits its triangle-inequality structure to learn goal-conditioned values. Contrastive reinforcement learning~\citep[CRL;][]{eysenbach2022contrastive} estimates a Monte Carlo goal-conditioned value function through contrastive learning and performs one-step policy improvement. Hierarchical implicit Q-learning~\citep[HIQL;][]{park2023hiql} extracts a two-level hierarchical policy from a single GCIVL value function. Its high-level policy predicts a latent representation of an intermediate \(m\)-step subgoal, while its low-level policy predicts primitive actions conditioned on that subgoal representation.

We additionally compare against option-aware temporally abstracted value learning~\citep[OTA;][]{ahn2026option}, which directly builds on HIQL and is the closest fixed-scale baseline to GITA. OTA replaces primitive one-step backups in the high-level value objective with temporally extended transitions using a fixed abstraction factor \(k\), thereby shortening the effective value-propagation horizon. Unlike GITA, however, OTA uses a single abstraction factor for all state-goal pairs within a task and therefore requires task-specific selection of \(k\). We use the task-specific abstraction factors and hyperparameters reported for OTA while otherwise following the corresponding HIQL/OGBench implementation.

\section{Additional Ablation Details}
\label{app:ablations}

We provide additional details on the ablations introduced in \Cref{sec:ablations}. Each variant changes a single component of GITA while leaving the remaining architecture, optimization procedure, goal sampling, and low-level policy unchanged. Unless stated otherwise, the abstraction-conditioned high-level value function is evaluated over the same candidate set $\mathcal{K}$ used by GITA. Recall that, for a training tuple $(s_t,s_{t+m},g)$, GITA computes
\begin{equation}
A^h(s_t,s_{t+m},g,k)
=
V_\theta^h(s_{t+m},g,k)
-
V_\theta^h(s_t,g,k)
\end{equation}
for every $k\in\mathcal{K}$ and uses the aggregated AWR weight
\begin{equation}
W_{\beta_h}
=
\frac{1}{|\mathcal{K}|}
\sum_{k\in\mathcal{K}}
\exp\left(
\beta_h A^h(s_t,s_{t+m},g,k)
\right)
\end{equation}
to train the shared high-level policy. The following ablations separately modify the value conditioning, the scale aggregation rule, or the candidate abstraction factors.

\paragraph{Unconditioned Value.}
This variant removes the abstraction factor from the high-level value function, replacing $V_\theta^h(s,g,k)$ with a single unconditioned value function $V_\theta^h(s,g)$. Value training still uses temporally abstract transitions spanning different numbers of primitive steps, but the realized abstraction factor is not provided to the value network. The resulting value function must therefore fit supervision from multiple temporal scales within the same mapping. Since scale-specific values are no longer available, the high-level policy is trained from the single advantage
\begin{equation}
A^h(s_t,s_{t+m},g)
=
V_\theta^h(s_{t+m},g)
-
V_\theta^h(s_t,g),
\end{equation}
using the standard AWR weight
\begin{equation}
\exp\left(
\beta_h A^h(s_t,s_{t+m},g)
\right).
\end{equation}
This ablation isolates whether training on multiple temporal scales is sufficient by itself, or whether explicitly representing the abstraction factor in the value function is necessary.

\paragraph{Highest Advantage Factor.}
This variant retains the abstraction-conditioned value function and computes $A^h(s_t,s_{t+m},g,k)$ for every $k\in\mathcal{K}$ as in GITA, but replaces soft aggregation across scales with hard selection. For each training sample, we select
\begin{equation}
k^\star
=
\arg\max_{k\in\mathcal{K}}
A^h(s_t,s_{t+m},g,k)
\end{equation}
and train the high-level policy using only
\begin{equation}
W_{\beta_h}^{\mathrm{max}}
=
\exp\left(
\beta_h
A^h(s_t,s_{t+m},g,k^\star)
\right).
\end{equation}
All other factors are discarded for that sample. This ablation tests whether the multi-scale policy update benefits from combining supervision across several temporal resolutions, or whether selecting only the scale with the largest advantage is sufficient.

\paragraph{Aggregate Advantage Before Exp.}
This variant preserves all scale-specific advantages but changes the order of aggregation and the nonlinear AWR transformation. GITA first exponentiates each advantage and then averages the resulting weights,
\begin{equation}
W_{\beta_h}
=
\frac{1}{|\mathcal{K}|}
\sum_{k\in\mathcal{K}}
\exp\left(
\beta_h A^h(s_t,s_{t+m},g,k)
\right).
\end{equation}
The ablation instead first averages the raw advantages,
\begin{equation}
\bar{A}^h(s_t,s_{t+m},g)
=
\frac{1}{|\mathcal{K}|}
\sum_{k\in\mathcal{K}}
A^h(s_t,s_{t+m},g,k),
\end{equation}
and then applies the AWR exponential,
\begin{equation}
W_{\beta_h}^{\mathrm{pre}}
=
\exp\left(
\beta_h
\bar{A}^h(s_t,s_{t+m},g)
\right).
\end{equation}
This removes the scale-specific amplification induced by applying the exponential before aggregation. The comparison therefore isolates whether preserving differences in the magnitude of the individual scale-conditioned advantages through the nonlinear AWR weighting is important for policy supervision.

\paragraph{GITA with Geometric $k$.}
The default GITA policy update evaluates each training sample on a fixed candidate set $\mathcal{K}$. This ablation replaces that set with freshly sampled abstraction factors at every high-level policy update. We sample the same number $K=|\mathcal{K}|$ of factors independently according to
\begin{equation}
k_j
\sim
\operatorname{Geom}(1-\zeta),
\qquad
p(k_j)
=
(1-\zeta)\zeta^{k_j-1},
\qquad
j=1,\ldots,K,
\end{equation}
with $\zeta=0.9$, corresponding to a mean abstraction factor of $10$. The parameter $\zeta$ is specific to this ablation and is distinct from the value-learning parameter $\alpha=\gamma$. The sampled set
\begin{equation}
\mathcal{K}_{\mathrm{geo}}
=
\{k_1,\ldots,k_K\}
\end{equation}
then replaces the fixed candidate set in the standard GITA aggregation,
\begin{equation}
W_{\beta_h}^{\mathrm{geo}}
=
\frac{1}{K}
\sum_{k\in\mathcal{K}_{\mathrm{geo}}}
\exp\left(
\beta_h A^h(s_t,s_{t+m},g;k)
\right).
\end{equation}
This change affects only the candidate factors used to construct the high-level policy supervision; the abstraction-conditioned value-learning procedure, including its sampling distribution parameterized by $\alpha$, is unchanged. The ablation tests whether GITA depends on a manually specified finite set of candidate scales or can obtain similar behavior from stochastic coverage of the temporal-scale space.

\section{Further Results and Visualizations}
\label{app:further_results}

\subsection{Scale-Dependent Policy Weights Across State--Goal Distances}
\label{app:scale_dependent_policy_weights}

To further visualize how different temporal scales contribute to high-level policy learning, we examine the scale-specific AWR weights as a function of state--goal distance. For each abstraction factor \(k\), we compute
\begin{equation}
w_k
=
\min\left\{
\exp\left(\beta_h A^h(s_t,s_{t+m},g;k)\right),
100
\right\},
\label{eq:app_scale_specific_awr_weight}
\end{equation}
where \(A^h(s_t,s_{t+m},g;k)\) is the abstraction-conditioned high-level advantage from \Cref{eq:geo_high_advantage}. The clipping threshold of \(100\) is also applied during training, so the plotted quantity corresponds to the final AWR weight used in the high-level policy update, up to the particular set of \(k\) values shown. This clipping convention is inherited from the official HIQL and OTA implementations. We group samples according to their geodesic state--goal distance and report the mean of \(w_k\) within each distance bin for each abstraction factor. Curves are averaged across seeds, with shaded regions indicating one standard deviation.

\begin{figure*}[t]
    \centering

    \includegraphics[width=\textwidth]{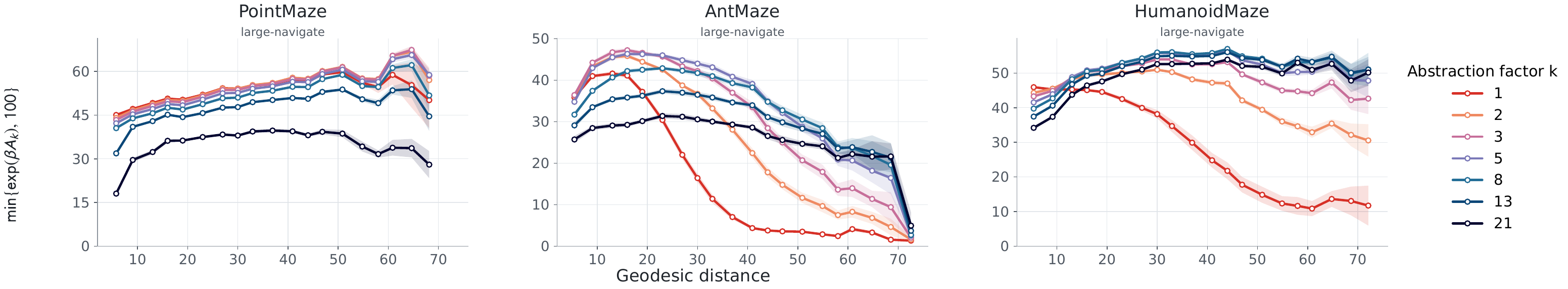}

    \vspace{0.15cm}

    \includegraphics[width=\textwidth]{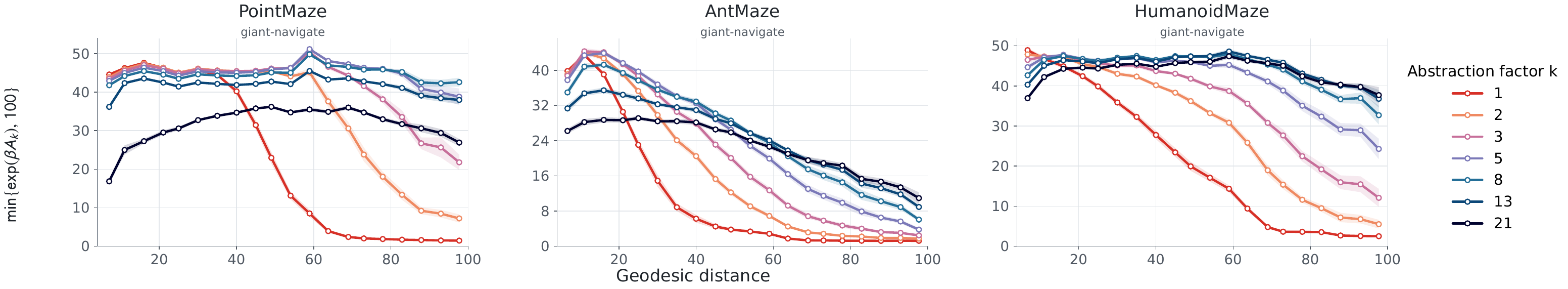}

    \vspace{0.15cm}

    \includegraphics[width=\textwidth]{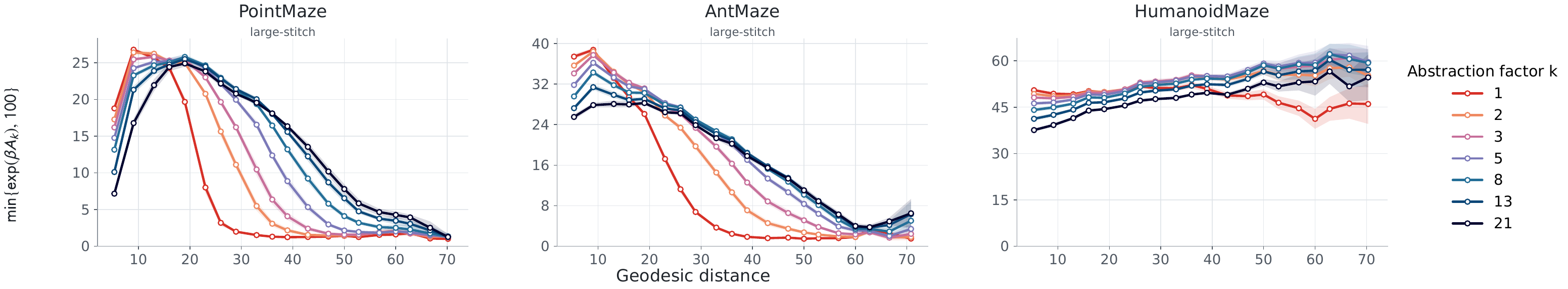}

    \vspace{0.15cm}

    \includegraphics[width=\textwidth]{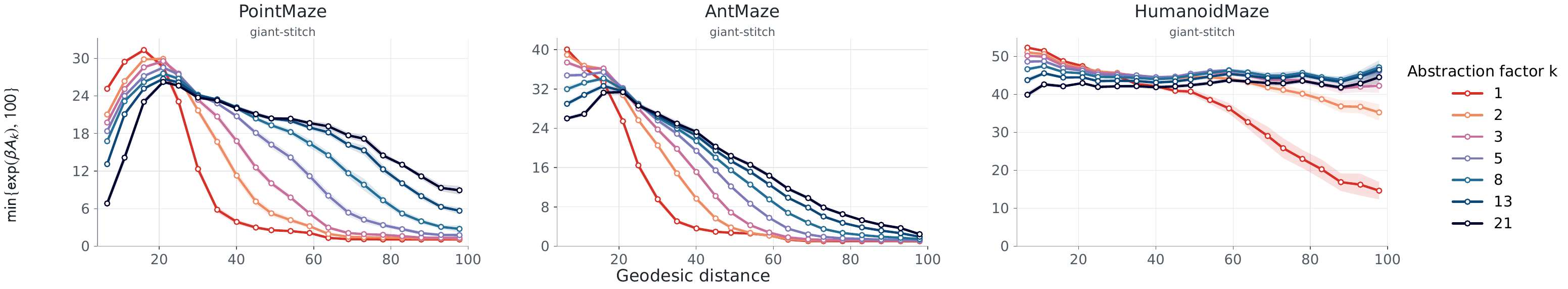}

    \caption{
        \textbf{Scale-specific AWR weights across navigation and stitching tasks.}
        Mean clipped exponentiated high-level advantage,
        \(\min\{\exp(\beta_h A^h),100\}\), as a function of geodesic state--goal distance for different abstraction factors \(k\).
        From top to bottom: \texttt{large-navigate}, \texttt{giant-navigate}, \texttt{large-stitch}, and \texttt{giant-stitch}.
        Curves show the mean across seeds and shaded regions indicate one standard deviation.
        Each plot contains the corresponding maze morphologies.
    }
    \label{fig:app_exp_advantage}
\end{figure*}

Recall from \Cref{eq:geo_multiscale_weight} that GITA averages the exponentiated advantages across \(k\). Thus, scales assigning larger positive advantages to a training tuple contribute disproportionately to its policy weight, without requiring an explicit scale-selection rule. Figure~\ref{fig:app_exp_advantage} shows this behavior across PointMaze, AntMaze, and HumanoidMaze for the \texttt{navigate} and \texttt{stitch} datasets and the \texttt{large} and \texttt{giant} maze layouts.

Across the maze environments, the scale-specific AWR weights exhibit a clear dependence on state--goal distance. Smaller abstraction factors lose their advantage signal more rapidly as distance increases. Larger factors retain informative weights over longer ranges, with the transition occurring progressively later as \(k\) increases. This pattern is especially clear in the \texttt{giant} layouts and is consistent with the reachability--resolution trade-off predicted by our analysis.

\subsection{Results Across Discount Factors}
\label{app:discount_factor_results}

To characterize the sensitivity of GITA to the discount factor, we report performance for each value considered in our sweep, \(\gamma\in\{0.98,0.99,0.995\}\), on the maze tasks in \Cref{tab:gita_discount_results}. Unlike the results reported in the main evaluation, this table uses only the final checkpoint after \(10^6\) gradient steps. Results are reported as mean \(\pm\) standard deviation over 8 seeds.

Because we set \(\alpha=\gamma\), this sweep changes both the discount factor and the distribution of abstraction factors. In particular, \(\gamma=\alpha=0.995\) gives \(\mathbb{E}[k]=200\), which matches the data episode length of PointMaze and AntMaze \texttt{stitch}. As a result, many sampled abstraction factors are truncated by trajectory boundaries, so part of the performance drop at larger \(\gamma\) may come from the induced geometric distribution rather than from discounting alone. This interaction further justifies re-tuning $\gamma$ for GITA.

\begin{table*}[t]
    \centering
    \caption{\textbf{Success rate (\%) of GITA for each discount factor considered in our sweep.} Results use only the final checkpoint after \(10^6\) gradient steps and are reported as mean \(\pm\) standard deviation over 8 seeds. The \emph{Best \(\gamma\) per Task} column reports the best task-level result together with the corresponding discount factor in parentheses. The best global discount factor is \(\gamma=0.99\), selected by the average performance across all tasks; the last column reports the corresponding task-level performance.}
    \label{tab:gita_discount_results}

    \footnotesize
    \setlength{\tabcolsep}{2.2pt}
    \renewcommand{\arraystretch}{1.15}

    \begin{tabular}{
        @{}
        l
        l
        l
        c
        c
        c
        c
        c
        @{}
    }
        \toprule

        \textbf{Environment}
        & \textbf{Type}
        & \textbf{Size}
        & \(\boldsymbol{\gamma=0.98}\)
        & \(\boldsymbol{\gamma=0.99}\)
        & \(\boldsymbol{\gamma=0.995}\)
        & \shortstack{\textbf{Best \(\boldsymbol{\gamma}\)}\\\textbf{per Task}}
        & \shortstack{\textbf{Best Global}\\\(\boldsymbol{\gamma=0.99}\)} \\

        \midrule


        \multirow{5}{*}{AntMaze}
        & navigate
        & large
        & \resultcell{91.0}{1.6}
        & \resultcell{89.0}{1.9}
        & \resultcell{0.7}{0.8}
        & \bestgammacell{91.0}{1.6}{0.98}
        & \resultcell{89.0}{1.9} \\

        & navigate
        & giant
        & \resultcell{78.8}{2.7}
        & \resultcell{65.8}{2.6}
        & \resultcell{2.2}{3.0}
        & \bestgammacell{78.8}{2.7}{0.98}
        & \resultcell{65.8}{2.6} \\

        \cmidrule(lr){2-8}

        & stitch
        & large
        & \resultcell{85.6}{2.9}
        & \resultcell{60.2}{27.8}
        & \resultcell{0.1}{0.4}
        & \bestgammacell{85.6}{2.9}{0.98}
        & \resultcell{60.2}{27.8} \\

        & stitch
        & giant
        & \resultcell{41.5}{5.8}
        & \resultcell{16.6}{13.2}
        & \resultcell{0.0}{0.0}
        & \bestgammacell{41.5}{5.8}{0.98}
        & \resultcell{16.6}{13.2} \\

        \cmidrule(lr){2-8}

        & explore
        & large
        & \resultcell{65.9}{12.5}
        & \resultcell{87.0}{8.8}
        & \resultcell{50.8}{21.9}
        & \bestgammacell{87.0}{8.8}{0.99}
        & \resultcell{87.0}{8.8} \\

        \midrule


        \multirow{4}{*}{HumanoidMaze}
        & navigate
        & large
        & \resultcell{50.8}{7.1}
        & \resultcell{72.2}{4.6}
        & \resultcell{68.2}{2.8}
        & \bestgammacell{72.2}{4.6}{0.99}
        & \resultcell{72.2}{4.6} \\

        & navigate
        & giant
        & \resultcell{54.1}{7.1}
        & \resultcell{89.6}{4.3}
        & \resultcell{83.2}{3.4}
        & \bestgammacell{89.6}{4.3}{0.99}
        & \resultcell{89.6}{4.3} \\

        \cmidrule(lr){2-8}

        & stitch
        & large
        & \resultcell{28.2}{4.2}
        & \resultcell{47.8}{4.5}
        & \resultcell{49.0}{2.3}
        & \bestgammacell{49.0}{2.3}{0.995}
        & \resultcell{47.8}{4.5} \\

        & stitch
        & giant
        & \resultcell{10.0}{1.3}
        & \resultcell{61.2}{6.5}
        & \resultcell{67.5}{2.9}
        & \bestgammacell{67.5}{2.9}{0.995}
        & \resultcell{61.2}{6.5} \\

        \midrule


        \multirow{4}{*}{PointMaze}
        & navigate
        & large
        & \resultcell{88.0}{10.4}
        & \resultcell{92.0}{8.6}
        & \resultcell{92.0}{5.9}
        & \bestgammacell{92.0}{5.9}{0.995}
        & \resultcell{92.0}{8.6} \\

        & navigate
        & giant
        & \resultcell{74.9}{12.7}
        & \resultcell{82.4}{4.7}
        & \resultcell{81.5}{7.7}
        & \bestgammacell{82.4}{4.7}{0.99}
        & \resultcell{82.4}{4.7} \\

        \cmidrule(lr){2-8}

        & stitch
        & large
        & \resultcell{72.7}{16.6}
        & \resultcell{61.5}{17.0}
        & \resultcell{20.6}{8.5}
        & \bestgammacell{72.7}{16.6}{0.98}
        & \resultcell{61.5}{17.0} \\

        & stitch
        & giant
        & \resultcell{10.3}{8.0}
        & \resultcell{56.2}{12.6}
        & \resultcell{34.4}{11.6}
        & \bestgammacell{56.2}{12.6}{0.99}
        & \resultcell{56.2}{12.6} \\

        \bottomrule
    \end{tabular}
\end{table*}

\end{document}